\documentclass[journal,comsoc]{IEEEtran}
\usepackage{graphicx}
\usepackage{amsmath,amsthm}
\usepackage{amssymb}
\usepackage{mathrsfs} 
\usepackage{enumerate}
\usepackage{multirow}
\usepackage{tabularx}
\usepackage{float}
\usepackage{color}
\usepackage{times}
\usepackage[table]{xcolor}

\usepackage{hyperref}
\hypersetup{
    colorlinks=true,
    linkcolor=blue,
   citecolor=blue,
    filecolor=magenta,      
    urlcolor=blue,
}

\def\R{\mathbb{R}}
\def\N{\mathbb{N}}
\def\S{\mathbb{S}}

\def\Rad{\mathscr{R}}

\def\K{\mathcal{K}}
\def\O{\mathcal{O}}

\definecolor{background}{rgb}{0.9373,0.9059,0.8078}
\definecolor{backgroundd}{rgb}{0.8836,0.8523,0.7547}
\begin{document}
\title{ Transport-based analysis, modeling, and learning from signal and data distributions}
\author{Soheil~Kolouri,~Serim Park,~Matthew~Thorpe,~Dejan~Slep\v{c}ev,~Gustavo~K.~Rohde}


\maketitle

\begin{abstract}
Transport-based techniques for signal and data analysis have received increased attention recently. Given their abilities to provide accurate generative models for signal intensities and other data distributions, they have been used in a variety of applications including content-based retrieval, cancer detection, image super-resolution, and statistical machine learning, to name a few, and shown to produce state of the art in several applications. Moreover, the geometric characteristics of transport-related metrics have inspired new kinds of algorithms for interpreting the meaning of data distributions. Here we provide an overview of the mathematical underpinnings of mass transport-related methods, including numerical implementation, as well as a review, with demonstrations, of several applications. Software accompanying this tutorial is available at \cite{Codes}. 
\end{abstract}

\IEEEpeerreviewmaketitle

\section{Introduction}
\subsection{\bf Motivation and goals}

Numerous applications in science and technology depend on effective modeling and information extraction from signal and image data. Examples include being able to distinguish between benign and malignant tumors from medical images, learning models (e.g. dictionaries) for solving inverse problems, identifying people from images of faces, voice profiles, or fingerprints, and many others.  Techniques based on the mathematics of optimal mass transport have received significant attention recently (see Figure \ref{fig:OTRef} and section \ref{sec:history}) given their ability to incorporate spatial (in addition to intensity) information when comparing signals, images, other data sources, thus giving rise to different geometric interpretation of data distributions. These techniques have been used to simplify and augment the accuracy of numerous pattern recognition-related problems. Some examples covered in this tutorial include image retrieval \cite{rubner2000earth,pele2009fast,li2013novel} , signal and image representation \cite{wang2013linear,park2015cumulative,seguy2015principal,kolouri2016radon,kolouri2016continuous}, inverse problems \cite{rabin2011wasserstein,swoboda2013convex,lellmann2014imaging,brauer2015cartoon}, cancer detection \cite{wang2011optimal,basu2014detecting,ozolek2014accurate,tosun2015detection}, texture and color modeling \cite{delon2004midway,rabin2011wasserstein,rabin2012wasserstein,ferradans2013static,rabin2014adaptive,ferradans2014regularized}, shape and image registration \cite{haker2001mass,haker2003monge,haker2004optimal,zhu2007image,ur20093d,museyko2009application,makihara2010earth,flamary2014optimal,lai2014multi,su2015optimal}, segmentation \cite{ni2009local,schmitzer2013object,rabin2015convex}, watermarking \cite{mathon2009optimization,mathon2014optimal}, and machine learning \cite{solomon2014wasserstein,courty2014domain,frogner2015learning,kolouri2016sliced,montavon2015wasserstein,el2012bayesian,kim2013efficient,reich2013nonparametric,oliver2014minimization,ramdas2015wasserstein}, to name a few. This tutorial is meant to serve as an \emph{introductory guide} to those wishing to familiarize themselves with these emerging techniques. Specifically we

\begin{itemize}
	\item provide a brief overview on the mathematics of optimal mass transport
	\item describe recent advances in transport related methodology and theory
	\item provide a pratical overview of their applications in modern signal analysis, modeling, and learning problems.
\end{itemize}
\emph{Software} accompanying this tutorial is available at \cite{Codes}.

\subsection{\bf Why transport?}

In recent years numerous techniques for signal and image analysis have been developed to address important learning and estimation problems. Researchers working to find solutions to these problems have found it necessary to develop techniques to compare signal intensities across different signal/image coordinates. A common problem in medical imaging, for example, is the analysis of magnetic resonance images with the goal of learning brain morphology differences between healthy and diseased populations. Decades of research in this area have culminated with techniques such as voxel and deformation-based morphology \cite{ashburner2000voxel,ashburner1998identifying} which make use of nonlinear registration methods to understand differences in tissue density and locations \cite{grenander1998computational,sotiras2013deformable}. Likewise, the development of dynamic time warping techniques was necessary to enable the comparison of time series data more meaningfully, without confounds from commonly encountered variations in time \cite{keogh2005exact}. Finally, researchers desiring to create realistic models of facial appearance have long understood that appearance models for eyes, lips, nose, etc. are significantly different and must thus be dependent on position relative to a fixed anatomy \cite{cootes2001active}. The pervasive success of these, as well as other techniques such as  optical flow \cite{baker2011database}, level-set methods \cite{cremers2007review}, deep neural networks \cite{schmidhuber2015deep}, for example, have thus taught us that 1) nonlinearity and 2) modeling the location of pixel intensities are essential concepts to keep in mind when solving modern regression problems related to estimation and classification. 

We note that the methodology developed above for modeling appearance and learning morphology, time series analysis and predictive modeling, deep neural networks for classification of sensor data, etc., is algorithmic in nature. The transport-related techniques reviewed below are nonlinear methods that, unlike linear methods such as Fourier, wavelets, and dictionary models, for example, explicitly model jointly signal intensities as well as their locations. Furthermore, they are often based on the theory of optimal mass transport from which fundamental principles can be put to use. Thus they hold the promise to ultimately play a significant role in the development of a theoretical foundation for certain subclasses of modern learning and estimation problems.

\subsection{\bf Overview and outline}

As detailed below in section \ref{sec:history} the optimal mass transport problem first arose due to Monge \cite{monge1781memoire}. It was later expanded by Kantarovich \cite{kantorovich1942translation,kantorovich1948monge,kantorovitch1958translocation,kantorovich1960mathematical} and found applications in operations research and economics. Section \ref{sec:formulation} provides an overview of the mathematical principles and formulation of optimal transport-related metrics, their geometric interpretation, and related embedding methods and signal transforms. We also explain Brenier's theorem \cite{brenier1991polar}, which helped pave the way for several practical numerical implementation algorithms, which are then explained in detail in section \ref{sec:numerics}. Finally, in section \ref{sec:applications} we review and demonstrate the application of transport-based techniques to numerous problems including: image retrieval, registration and morphing, color and texture analysis, image denoising and restoration, morphometry, super resolution, and machine learning. As mentioned above, software implementing the examples shown can be downloaded from \cite{Codes}.

\begin{figure}
\centering
\includegraphics[width=\columnwidth]{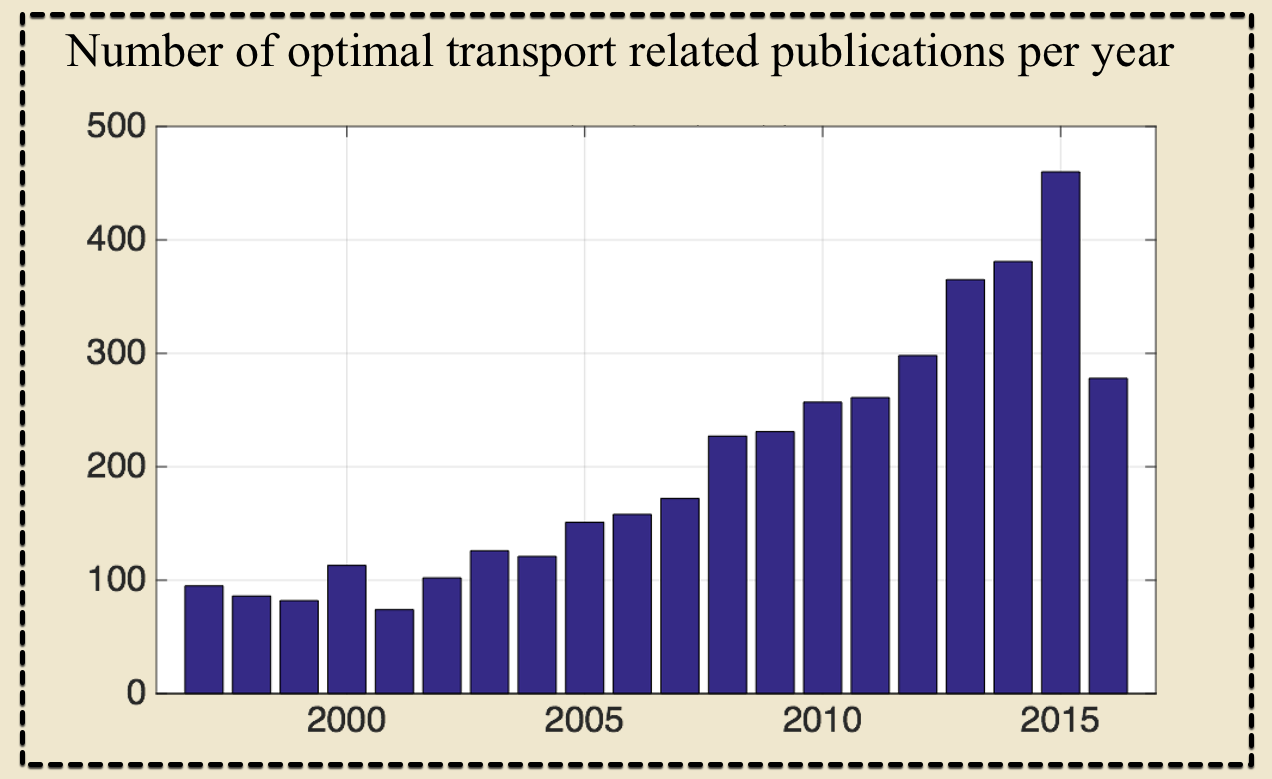}
\caption{The number of publications by year that contain any of the keywords: `optimal transport', `Wasserstein', `Monge', `Kantorovich', `earth mover' according to \url{webofknowledge.com}. By comparison there were 819 in total pre 1997. The 2016 statistics is correct as of 29th August 2016.}
\label{fig:OTRef}
\end{figure}

\section{A brief historical note}
\label{sec:history}

 The optimal mass transport problem seeks the most efficient way of transforming one distribution of mass to another, relative to a given cost function. The problem was initially studied by the French Mathematician Gaspard Monge in his seminal work ``M\'{e}moire sur la th\'{e}orie des d\'{e}blais et des remblais''  \cite{monge1781memoire} in 1781. Between 1781 and 1942, several mathematicians made valuable contributions to the mathematics of the optimal mass transport problem among which are Arthur Cayley \cite{cayley1883monge}, Charles Duplin \cite{dupinapplications}, and Paul Appell \cite{appell1928probleme}.  In 1942, Leonid V. Kantorovich, who at that time was unaware of Monge's work, proposed a general formulation of the problem by considering optimal mass transport plans, which as opposed to Monge's formulation allows for mass splitting \cite{kantorovich1942translation}. Six years later and in 1948, Kantorovich became aware of Monge's work and made the connection between his work and that of the French mathematician in his short paper, ``A Problem of Monge'' \cite{kantorovich1948monge}. Kantorovich shared the 1975 Nobel Prize in Economic Sciences with Tjalling Koopmans for his work in the optimal allocation of scarce resources.  Kantorovich's contribution is considered as ``the birth of the modern formulation of optimal transport'' \cite{villani2008optimal} and it made the optimal mass transport problem an active  field of research in the following years \cite{russell1969letters, levin1977duality, knott1984optimal, rachev1985monge}. Among the published works in this era is the prominent work of Sudakhov \cite{sudakov1979geometric} in 1979 on the existence of the optimal transport map. Later on, a gap was found in Sudakhov's work \cite{sudakov1979geometric} which was  
 eventually filled by Ambrosio in 2003 \cite{ambrosio2003lecture}.
 
A significant portion of the theory of the optimal mass transport  problem was developed in the Nineties. Starting with Brenier's seminal work on characterization, existence, and uniqueness of optimal transport maps \cite{brenier1991polar}, followed by Cafarrelli's work on regularity conditions of such mappings \cite{caffarelli1992regularity}, Gangbo and McCann's work on geometric interpretation of the problem \cite{gangbo1995optimal, gangbo1996geometry}, and Evans and Gangbo's work on formulating the problem through differential equations and specifically the p-Laplacian equation \cite{evans1997partial, evans1999differential}. A more thorough history and background on the optimal mass transport problem can be found in Rachev and R\"{u}schendorf's book ``Mass Transportation Problems'' \cite{rachev1998mass}, Villani's book ``Optimal Transport: Old and New" \cite{villani2008optimal}, and Santambrogio's book `` Optimal transport for applied mathematicians'' \cite{santambrogio2015optimal}, or in shorter surveys such as that of Bogachev and Kolesnikov's \cite{bogachev2012monge} and Vershik's \cite{vershik2013long}.
 
The significant contributions in mathematical foundations of the optimal transport problem together with recent advancements in numerical methods \cite{cuturi2013sinkhorn, benamou2015iterative, benamou2000computational,oberman2015efficient, urrehman07multigrid} have spurred the recent development of numerous data analysis techniques for modern estimation and detection (e.g. classification) problems. Figure \ref{fig:OTRef} plots the number of papers related to the topic of optimal transport that can be found in the public domain per year demonstrating significant growth in these techniques.

\section{Formulation of the problem and methodology}
\label{sec:formulation}

In this section we first review the two classical formulations of the optimal transport problem (i.e. Monge's and Kantorovich's formulations). Next, we review the geometrical characteristics of the problem, and finally review the transport based signal/image embeddings.  

\subsection{\bf Optimal Transport: Formulation}

\subsubsection{\textit{\textbf{  Monge's formulation}}}

The Monge optimal mass transportation problem is formulated as follows. Consider two probability measures $\mu$ and $\nu$ defined on measure spaces $X$ and $Y$.
In most applications $X$ and $Y$ are subsets of $\R^d$ and
$\mu$ and $\nu$ have
 density functions which we denote by $I_0$ and $I_1$, $d\mu(x)=I_0(x)dx$ and $d\nu(x)=I_1(x)dx$, (originally representing the height of a pile of soil/sand and the depth of an excavation). 
 Monge's optimal transportation problem is to
  find a measurable map $f:X\rightarrow Y$ that pushes $\mu$ onto $\nu$ and minimizes the following objective function, 
\begin{eqnarray}	
M(\mu,\nu)=\inf_{ f\in MP} \int_X c(x,f(x))d\mu(x)
\label{eq:Monge}
\end{eqnarray}
where $c:X\times Y\rightarrow \R^+$ is the cost functional, and $MP:= \{ f:X\rightarrow Y~|~ f_{\#}\mu=\nu \}$ where $f_{\#}\mu$ represents the pushforward of measure $\mu$ and is characterized as,  
\begin{eqnarray}
\int_{f^{-1}(A)} d\mu(x) =\int_{A} d\nu(y) ~~ \text{for any measurable } A\subset Y.
\label{eq:MPmap}
\end{eqnarray}

\begin{figure}[t]
\centering
\includegraphics[width=\columnwidth]{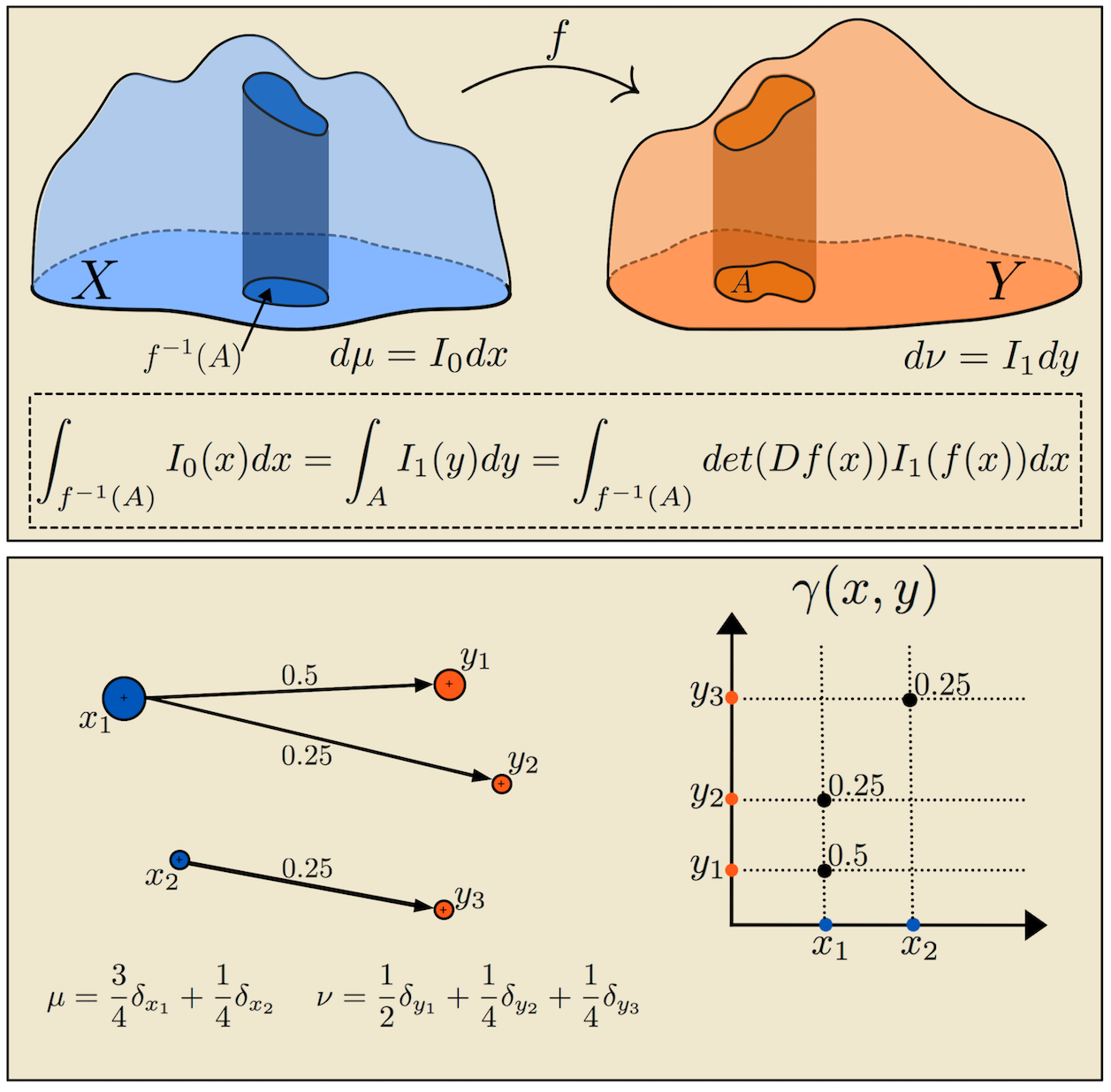}
\caption{Monge transport map (top panel) and Kantorovich's transport plan (bottom panel).}
\label{fig:MK}
\end{figure}

If  $\mu$ and $\nu$ have densities and $f$ is smooth and one to one then the equation above can be written in a differential form as 
\[ det(Df(x))I_1(f(x))=I_0(x) \] almost everywhere, where $Df$ is the Jacobian of $f$ (See Figure \ref{fig:MK}, top panel). Such measurable maps $f\in MP$ are sometimes called `transport maps' or `mass preserving maps'. Simply put, the Monge formulation of the problem seeks the best pushforward map that rearranges measure $\mu$ into measure $\nu$ while minimizing a specific cost function. Monge considered the Euclidean distance as the cost function in his original formulation, $c(x,f(x))=|x-f(x)|$. Note that both the objective function and the constraint in Equation \eqref{eq:Monge} are nonlinear with respect to $f(x)$. Hence, for over a century the answers to questions regarding existence and characterization of the Monge's problem remained unknown. 

It should be mentioned that, for certain measures the Monge's formulation of the optimal transport problem is ill-posed; in the sense that there is no transport map to rearrange one measure to another. For instance, consider the case where $\mu$ is a Dirac mass while $\nu$ is not. 
Kantorovich's formulation alleviates this problem by finding the optimal transport plan as opposed to the transport map.  

\subsubsection{\textit{\textbf{  Kantorovich's formulation}}}
Kantorovich formulated the transportation problem by optimizing over transportation plans, where a transport plan is a probability measure $\gamma \in P(X\times Y)$ with marginals $\mu$ and $\nu$. One can think of $\gamma$ as the joint distribution of $I_0$ and $I_1$ describing how much `mass' is being moved to different coordinates. That is let $A$ be a measurable subset of  $X$ and similarly $B \subseteq Y$. The quantity $\gamma(A,B)$ tells us how much `mass' in set $A$ is being moved to set $B$. Now let $\gamma\in P(X\times Y)$ be a plan with marginals $\mu$ and $\nu$, i.e., 
\begin{eqnarray*}
(\Gamma_X)_\# \gamma=\mu ~~~ \text{and}~~~ (\Gamma_Y)_\# \gamma=\nu
\end{eqnarray*}
where $\Gamma_X:X\times Y \rightarrow X$ and $\Gamma_Y:X\times Y \rightarrow Y$ are the canonical projections. Let $\Gamma(\mu,\nu)$ be the set of all such plans. Kantorovich's formulation can then be written as, 
\begin{eqnarray}
K(\mu,\nu)=\min_{\gamma\in\Gamma(\mu,\nu)}  \int_{X\times Y}c(x,y)d\gamma(x,y).
\label{eq:Kantorovich}
\end{eqnarray}
The minimizer of the optimization problem above, $\gamma^*$, is called the optimal transport plan. Note that unlike the Monge problem, in Kantorovich's formulation the objective function and the constraints are linear with respect to $\gamma(x,y)$. Moreover, Kantorovich's formulation is in the form of a convex optimization problem. Note that the Monge problem is more restrictive than the Kantorovich problem. That is, in Monge's version, mass from a single location in one domain is being sent to a single location in the domain of the other measure. Kantorovich's formulation considers transport plans which can deal with arbitrary measurable sets and has the ability to distribute mass from the one location in one density to multiple locations in another (See Figure \ref{fig:MK}, bottom panel). 
For any transport map $f : X \rightarrow Y$ there is an associated transport plan, given by
\begin{equation} \label{eq:mapplan}
	\gamma=(\mathrm{Id}\times f)_{\#}\mu,
\end{equation}
which means that
$\gamma(A,B) = \mu \left (  \{ x \in A : f(x) \in B \}  \right)$, or in other words for any integrable function $h(x,y)$ on measure space $X \times Y$
\begin{eqnarray}
\int c(x,y) d\gamma(x,y) =\int c(x,f(x))d\mu(x).
\end{eqnarray}
Furthermore  when an optimal transport map $f^*$ exists, with the optimal transport then $\gamma^*$ given by \eqref{eq:mapplan} is an optimal transportation plan  \cite{villani2008optimal}.
 Note that the converse does not always hold.  Finally, the following relationship holds between Equations \eqref{eq:Monge} and \eqref{eq:Kantorovich} \cite{bogachev2012monge,villani2008optimal}, 
\begin{eqnarray}
K(\mu,\nu)\leq M(\mu,\nu).
\end{eqnarray}

The Kantorovich problem is especially interesting in a discrete setting, that is
 for probability measures  of the form $\mu=\sum_{i=1}^M p_i\delta_{x_i}$ and $\nu=\sum_{j=1}^N q_j \delta_{y_j}$, where $\delta_{x_i}$ is a dirac measure centered at $x_i$, the Kantorovich problem can be written as, 
\begin{eqnarray}
K(\mu,\nu)&=& \min_{\gamma} \sum_i\sum_j c(x_i,y_j)\gamma_{ij}\nonumber\\
&s.t.&~ \sum_j \gamma_{ij}= p_i,~\sum_i \gamma_{ij}= q_j\nonumber\\
&& \gamma_{ij}\geq 0 ,  i=1,...,M,~j=1,...,N 
\label{eq:lp}
\end{eqnarray}
where $\gamma_{ij}$ identifies how much of the mass particle $m_i$ at $x_i$ needs to be moved to $y_j$ (See Figure \ref{fig:MK}, bottom panel). Note that the optimization above has a linear objective function and linear constraints, therefore it is a linear programming problem. We note that the problem is convex, but not strictly so, and the constraint provides a polyhedral set of $M\times N$ matrices, $\Gamma(\mu,\nu)$.

In practice, a non-discrete measure is often approximated by a discrete measure and the Kantorovich problem is solved through the linear programming optimization expressed in equation \eqref{eq:lp}. An important result that allows this discretization is the following. If $\mu^h$ and $\nu^h$ are sequences of measures that weakly converge to $\mu$ and $\nu$ respectively and $\gamma^h$ is the optimal transport plan between $\mu^h$ and $\nu^h$
then, up to subsequences, $\gamma^h$ converges weakly to the optimal transport plan between $\mu$ and $\nu$, $\gamma^*$ (see for
example~\cite[Thm 5.20]{villani2008optimal}).
This result enables the application of discrete OT methods to more general measures.

Here we note that the Kantorovich problem can be extended to multi-marginal problems where instead of working in $X\times Y$ one defines $\chi:=X_1\times X_2 \times ...\times X_N$, the cost $c:\chi\rightarrow \R^+$, and marginals are defined as $(\Gamma_{X_i})_\#\gamma=\mu_i$. Multi-marginal transport problems and Martingale optimal transport problems \cite{pass2014multi,santambrogio2015optimal,colombo2013multimarginal}  are two variations of the transport problem, which have recently attracted attention in economics \cite{pass2014multi,hobson2015robust}.

\subsubsection{\textit{\textbf{  Dual formulation}}}
\label{sec:dual}

The focus of the Kantorovich problem is to minimize the {\it cost} of transportation. Villani \cite{villani2008optimal} describes the problem by an intuitive example using the transportation of bread between bakeries and caf\'{e}s. Consider a company that owns bakeries throughout a city, which produce loaves that should be transported to caf\'{e}s. If this company were to take delivery into its own hands, then it would like to minimize the transportation costs. Minimizing the cost of the transportation between bakeries and caf\'{e}s is the focus of Kantorovich's problem. Now imagine that a second company comes into play, which offers to take care of all transportation (maybe they have a few tricks to do the transportation in a very efficient manner) offering competitive prices for buying bread at bakeries and selling them at caf\'{e}s. Let $\phi(x)$ be the cost of bread bought at bakery $x$, and $\psi(y)$ be its selling price at caf\'{e} $y$. Before the second company showed up, the first company's selling price was equal to  $\phi(x)+c(x,y)$, which is the price of the bread plus the delivery fee (transportation cost). Now if the second company wants to be competitive, it needs to sell bread to caf\'{e}s with a price which is lower or equal to that which the first company offers. Meaning that, 
\begin{eqnarray}
\forall(x,y), ~~~\psi(y)\leq c(x,y)+\phi(x)
\end{eqnarray}
At the same time, the second company of course wants to maximize its {\it profit}, which can be formulated as, 
\begin{eqnarray}
KD(\mu,\nu)=&\sup_{\psi, \phi}& \int_Y \psi(y)d\nu(y)-\int_X \phi(x)d\mu(x)\nonumber\\
&s.t.&~\psi(y)-\phi(x)\leq c(x,y),~~\forall(x,y)
\label{eq:kdual}
\end{eqnarray}
Note that maximizing the profit is the dual problem \cite{kantorovich1960mathematical, rachev1998mass,villani2008optimal,santambrogio2015optimal,bogachev2012monge} of the Kantorovich formulation. In addition, since the primal problem is convex we have, 
\begin{eqnarray}
K(\mu,\nu)=KD(\mu,\nu).
\end{eqnarray}
It is apparent from this scenario that the optimal prices for the dual problem must be tight meaning that, the second company cannot increase the selling price, $\psi(y)$, or decrease the buying price, $\phi(x)$. Formally,
\begin{eqnarray}
\psi(y)&=& \operatorname{inf}_x (\phi(x)+c(x,y))\\ \phi(x)&=& \operatorname{sup}_y (\psi(y)-c(x,y)) \nonumber
\label{eq:cconjugate}
\end{eqnarray}
The equations above imply that the pair $\psi$ and $\phi$ are related through a {\it c-transform} (see \cite{villani2008optimal} for the definition and more detail), which coincides with the Legendre-Fenchel transform \cite{boyd2004convex} when $c(x,y)=-\langle x,y\rangle$. Finally, for the quadratic transport cost, $c(x,y)=\frac{1}{2}|x-y|^2$, and when an optimal transport map exists, the optimal arguments of the dual problem and the optimal transport map are related through, $f^*(x)=\nabla ( \frac12 |x|^2 - \phi^*(x))$.

\subsubsection{\textit{\textbf{Basic properties}}}
\label{sec:brenier}

The existence of a solution for the Kantorovich problem follows from 
 Ulham's  \cite{ulam1960collection} and Prokhorov's \cite{prokhorov1956convergence}  theorems. Ulham's theorem states that $\Gamma(\mu,\nu)$ is a tight set of  probability measures defined on $X\times Y$. On the other hand,  Prokhorov's theorem \cite{prokhorov1956convergence} states that $\Gamma(\mu,\nu)$ is relatively compact, and hence the limit of a sequence of  transport plans $(\gamma_n)\subset \Gamma(\mu,\nu)$ where $\gamma_n\rightarrow \gamma$ is also in $\Gamma(\mu,\nu)$. Using these theorems one can show the following theorem \cite{ambrosio2013user, villani2008optimal}: {\it For a lower semicontinuous and bounded from below cost function, $c(x,y)$ there exists a minimizer to the Kantorovich problem.}

In engineering applications the common cost functions almost always satisfy the existence of a transport plan. A further important question is regarding the existence of an optimal transport map instead of a plan. Brenier \cite{brenier1991polar} addressed this problem for the special case where $c(x,y)=|x-y|^2$. Bernier's results was later relaxed to more general cases by Gangbo and McCann \cite{gangbo1996geometry}, which led to the following theorem: 

{\bf Theorem} Let $\mu$ and $\nu$ be two Borel probability measures on compact measurable supports $X$ and $Y$, respectively. 
When $c(x,y)=h(x-y)$ for some strictly convex function $h$ and  $\mu$ is absolutely continuous with respect to the Lebesgue measure, then there exists a unique optimal transportation map $f^*:X\rightarrow Y$ such that $f^*_\#\mu=\nu$,
\begin{eqnarray}
\int_X h(x-f^*(x))d\mu(x)=\operatorname*{min}_{\gamma\in \Pi(\mu,\nu)}\int_{X\times Y} h(x-y)d\gamma(x,y).\nonumber\\
\end{eqnarray}
In addition, the optimal transport plan is unique, and thus equal to $\gamma(x,y)=(\mathrm{Id}\times f^*)_{\#}\mu$ (See Eq. \ref{eq:mapplan}). Moreover, $f^*$ is characterized by the gradient of a c-concave function $\phi:X\rightarrow \R$ as follows,
\begin{equation}
 f^*(x)=x-\nabla h^{-1}(\nabla\phi(x)).
 \end{equation}
 For a proof see \cite{ambrosio2013user, gangbo1996geometry, villani2008optimal}. Note that,  $h(x)=\frac{1}{2}|x|^2$ is rather a special case, because the gradient of $h$ is equal to identity,$\nabla h=\mathrm{Id}$, and the optimal transport map is simply characterized as $f^*(x)=x-\nabla\phi(x)$. 

\subsection{\bf Optimal Mass Transport: Geometric properties}

\subsubsection{\textit{\textbf{  Wasserstein metric}}}
Let $\Omega$ be a subset of $\R^d$ on which the measures we consider are defined. In most applications $\Omega$ is the domain where the signal is defined and thus bounded. 
Let $P_p(\Omega)$ be the set of Borel probability measures on $\Omega$, with finite $p$'th moment, that is
the set of probability measures $\mu$ on $\R^d$ such that $\int_\Omega |x|^p d \mu(x) < \infty$.
The p-Wasserstein metric, $W_p$, for $p\geq1$ on $P_p(\Omega)$ is then defined as using the optimal transportation problem \eqref{eq:Kantorovich} with the cost function $c(x,y)=|x-y|^p$. For $\mu$ and $\nu$ in $P_p(\Omega)$, 
\begin{eqnarray}
W_p(\mu,\nu)=\left(\operatorname*{inf}_{\gamma\in\Gamma(\mu,\nu)} \int_{\Omega\times \Omega} |x-y|^p d \gamma(x,y) \right)^{\frac{1}{p}}.
\end{eqnarray}
For any $p \geq 1$, $W_p$ is a metric on $P_p(\Omega)$.
The metric space $(P_p(\Omega),W_p)$ is referred to as the p-Wasserstein space. 
If $\Omega$ is bounded then  for any $p \geq 1$,  $W_p$ metrizes  the weak convergence of measures on $P(\Omega)$. That is the convergence with respect to $W_p$ is equivalent to weak convergence of measures. 

Note that the $p$-Wasserstein metric can equivalently be defined using the dual Kantorovich problem,
\begin{eqnarray}
W_p(\mu,\nu)=\left(\operatorname*{sup}_\phi \left\{ \int_{\Omega}\phi(x) d\mu(x)-\int_{\Omega}\phi^c(y) d\nu(y)\right\}\right)^{\frac{1}{p}}
\end{eqnarray}
where $\phi^c(y)= \operatorname{inf}_x \{ \phi(x)- |x-y|^p \}$.

For  the specific case of $p=1$ the p-Wasserstein metric is also known as the  Monge--Rubinstein \cite{villani2008optimal} metric, or the earth mover distance \cite{rubner2000earth}, which can also be expressed in the following form
\begin{eqnarray}
W_1(\mu,\nu)=\operatorname*{sup}_{Lip(\phi)\leq1} \left\{ \int_{\Omega}\phi(x) d(\mu(x)-\nu(x))\right\},
\end{eqnarray}
where $Lip(\phi)$ is the Lipschitz constant for $\phi$.

 The p-Wasserstein metric for one-dimensional probability measures is specifically interesting due to its simple and unique characterization. For 
  one-dimensional probability measures $\mu$ and $\nu$  on $\R$ 
the optimal transport map has a closed form solution.  Let $F_\mu$ be the cumulative distribution function of  a measure  $\mu \in P_p(\R)$
\begin{eqnarray}
F_\mu(x)=\mu((-\infty,x))
\end{eqnarray}
Note that this is a nondecreasing function going from $0$ to $1$, which is continuous if 
$\mu$ has a density. 
We define the \emph{pseudoinverse} of $F_\mu$ as follows: for $z \in (0,1)$, 
$F^{-1}(z)$ is the smallest $x$ for which $F_\mu(x) \geq z$, that is
\begin{equation} \label{pseudoinv}
F_\mu^{-1}(z) = \inf\, \{ x  \in \R \::\: F_\mu(x) \geq z \}
\end{equation}
If $\mu$ has positive density then $F_\mu$ is increasing (and thus invertible) and 
the inverse of the function $F_\mu$ is equal to the pseudoinverse we just defined. In other words the pseudoinverse is a generalization of the notion of the inverse of a function. 
The pseudoinverse (i.e. the inverse if the densities of $\mu$ and $\nu$ are positive)
 provides a closed form solution for the p-Wasserstein distance:
\begin{eqnarray}
W_p(\mu,\nu)=\left( \int_0^1 |F^{-1}_\mu(z) -F^{-1}_\nu(z)|^p dz \right)^\frac{1}{p}.
\end{eqnarray}
%
The closed-form solution of the p-Wasserstein distance in one dimension is an attractive property, as it alleviates the need for optimization. This property was employed in the Sliced Wasserstein metrics as defined below.

\begin{figure}[t]
\centering
\includegraphics[width=\columnwidth]{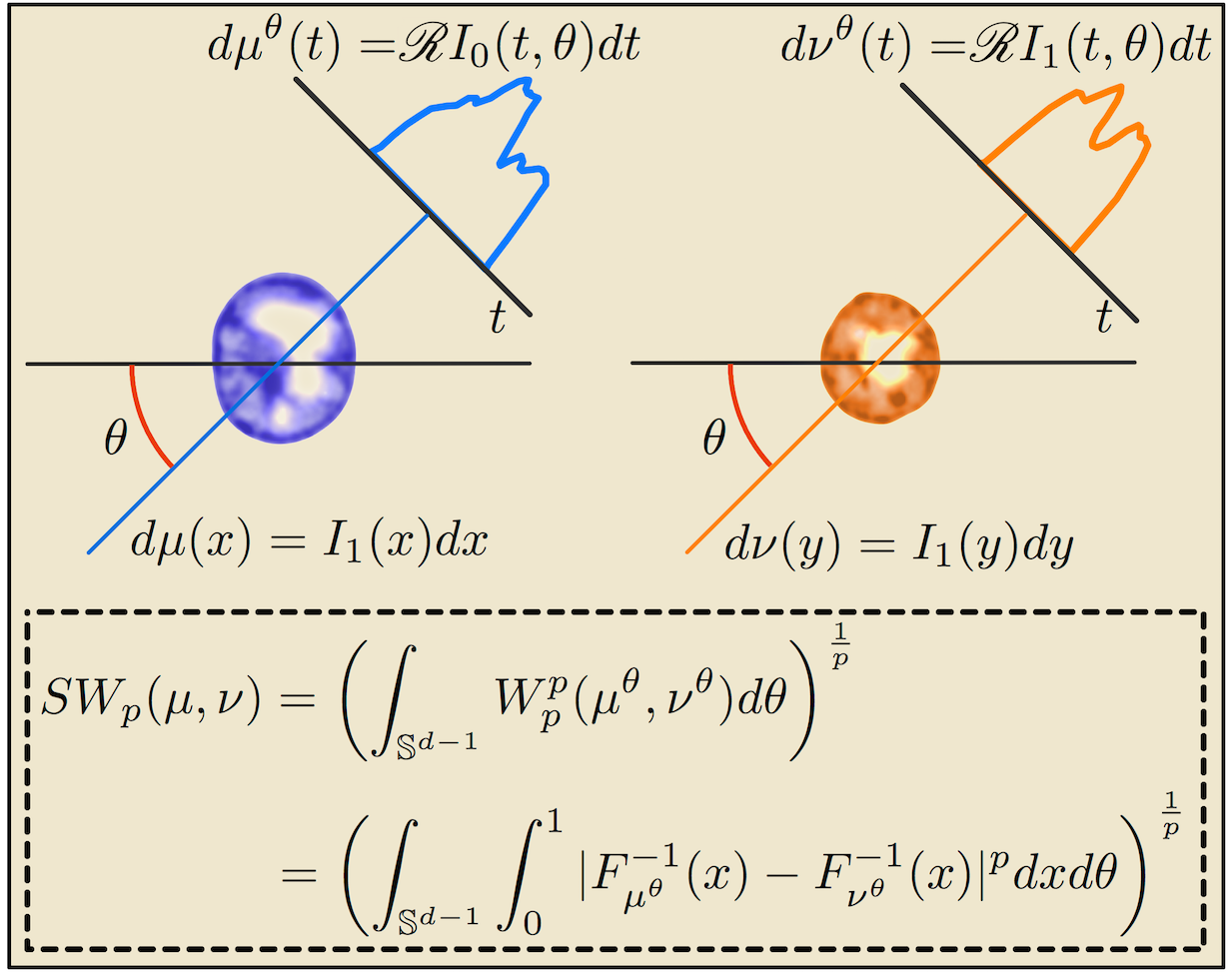}
\caption{Slicing measures $\mu$ and $\nu$ and the corresponding Sliced Wasserstein distance between them. The images represent segmented nuclei extracted from histopathology images.}
\label{fig:SW}
\end{figure}

\begin{figure*}[t!]
\centering
\includegraphics[width=\linewidth]{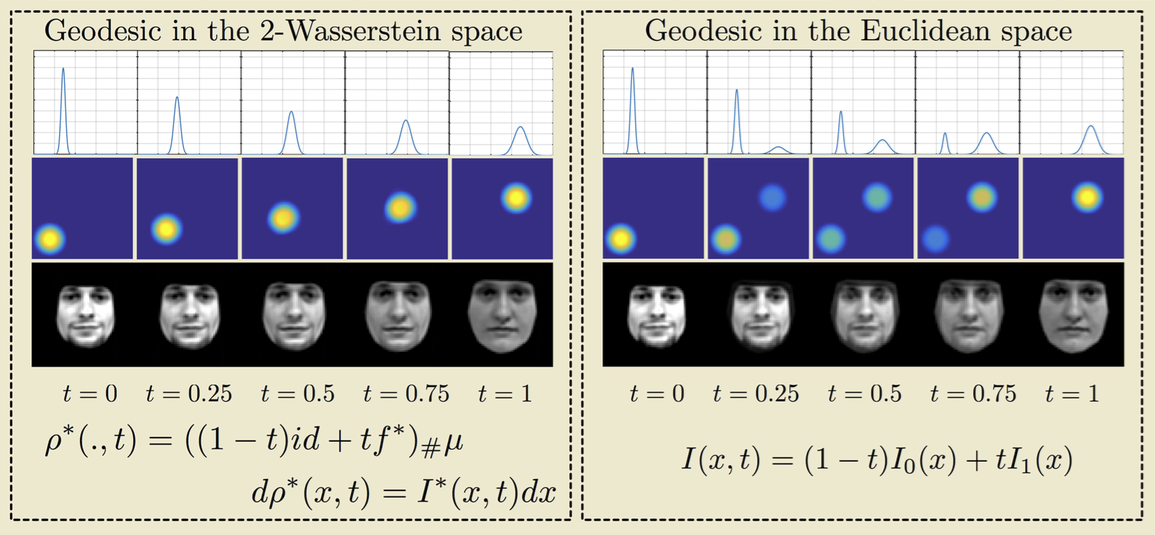}
\caption{Geodesic in the 2-Wasserstein space (left panel), and visualization of the geodesic, $\rho^*$, for one-dimensional signals and two-dimensional images (right panel). Note that the geodesic captures the nonlinear structure of the signals and images and provides a natural morphing.}
\label{fig:geodesic}
\end{figure*}

\subsubsection{\textit{\textbf{ Sliced-Wasserstein Metric}}}

 The idea behind the Sliced Wasserstein metric is to first obtain a family of one-dimensional representations for a higher-dimensional probability distribution through projections (slicing the measure), and then calculate the distance between two input distributions as a functional on the Wasserstein distance of their one-dimensional representations. In this sense, the distance is obtained by solving several one-dimensional optimal transport problems, which have closed-form solutions. 

Slicing a measure is closely related to the well known Radon transform in the imaging and image processing community \cite{bonneel2015sliced, kolouri2016radon}. The $d$-dimensional Radon transform $\Rad$ maps a function  $I\in L_1(\R^d)$ where $L_1(\R^d):=\{ I:\R^d \rightarrow \R | \int_{\R^d} |I(x)|dx \leq \infty\}$ into the set of its integrals over the hyperplanes of $\R^n$ and is defined as, 
\begin{eqnarray}
\Rad I(t,\theta):=\int_{\R} I(t\theta+s\theta^{\perp})ds,~~\forall t\in \R,~\forall \theta\in \S^{d-1}
\end{eqnarray}
here $\theta^{\perp}$ is the subspace orthogonal to $\theta$, and $\S^{d-1}$ is the unit sphere in $\R^{d}$. Note that $\Rad: L_1(\R^d)\rightarrow L_1(\R\times \S^{d-1})$. One can slice a probability measure $\mu$ defined on $\R\times \S^{d-1}$ into its conditional  measures with respect to the uniform measure on $\S^{d-1}$ to obtain a measure $\mu^\theta$, which satisfies,
\begin{eqnarray}
\int_{\R*\S^{d-1}} g(t,\theta)d\mu(t,\theta)=\int_{\S^{d-1}}\left( \int_{\R} g(t,\theta)d\mu^\theta(t)\right)d\theta.
\end{eqnarray} 
for $\forall g\in L_1(\R\times\S^{d-1})$. The Sliced Wasserstein metric, for continuous probability measures $\mu$ and $\nu$ on $\R^d$ is then defined as, 
\begin{eqnarray}
SW_p(\mu,\nu)=\left(\int_{\S^{d-1}} W^p_p(\mu^\theta,\nu^\theta) d\theta\right)^{\frac{1}{p}}
\end{eqnarray}
where $p\geq 1$, and $W_p$ is the Wasserstein metric, which for one dimensional measures $\mu^\theta$ and $\nu^\theta$ has a closed form solution (See Figure \ref{fig:SW}). For more details and definitions of the Sliced Wasserstein metric we refer the reader to \cite{rabin2012wasserstein,lai2014multi,bonneel2015sliced, kolouri2016radon}.

\subsubsection{\textit{\textbf{Wasserstein spaces, geodesics, and Riemannian structure}}}
\label{subsec:fluid}

In this section we assume that $\Omega$ is convex. 
Here we highlight that the p-Wasserstein space  $(P_p(\Omega),W_p)$ is not just a metric space, but has additional geometric structure. In particular for any $p\geq 1$ and any $\mu_0, \mu_1 \in P_p(\Omega)$ there exists a continuous path between $\mu_0$ and $\mu_1$ whose length is the distance between $\mu_0$ and $\mu_1$. 

Furthermore the space with $p=2$ is special as it possesses a structure of an formal, infinite dimensional, Riemannian manifold. That structure was first noted by Otto \cite{otto2001geometry} who developed the formal calculations for using this structure. The weak setting in which the notions of tangent space, Riemannian metric were made rigorous was developed by Ambrosio, Gigli and Savar\'e \cite{AGS}. A slightly less general, but more approachable introduction is available in  \cite{ambrosio2013user} and also in \cite{santambrogio2015optimal}. The notion of curvature was developed in \cite{Gigli2, lott2008some}.

Here we review the two main notions, which have a wide use.
Namely we characterize the geodesics in $(P_p(\Omega),W_p)$ and in the case $p=2$ describe what is the local, Riemannian metric of $(P_2(\Omega),W_2)$. Finally we 
state the seminal result of Benamou and Brenier \cite{benamou2000computational} who provided a characterization of geodesics via action minimization which is useful in computations and also gives an intuitive explanation of the Wasserstein metric.

We first recall the definition of the length of a curve in a metric space. Let $(X,d)$ be a metric space and $\mu : [a,b] \to X$. Then the length of $\mu$, denoted by $L(\mu)$ is
\begin{align*}  L(\mu)  =  \sup_{m \in \N,\, a=t_0 <t_1 < \cdots < t_m=b} \: \sum_{i=1}^m d(\mu(t_{i-1}), \mu(t_i)). 
\end{align*}
A metric space $(X,d)$ is a \emph{geodesic space} if for any $\mu_0$ and $\mu_1$ there exists a curve $\mu:[0,1] \to X$ such that $\mu(0)=\mu_0$, $\mu(1)= \mu_1$ and 
for all $0\leq s < t \leq 1$, $d(\mu(s), \mu(t)) = L(\mu|_{[s,t]})$. In particular the length of $\mu$ is equal to the distance from $\mu_0$ to $\mu_1$. 
Such curve $\mu$ is called a \emph{geodesic}. 
The existence of geodesic is useful as it allows one to define the average of $\mu_0$ and $\mu_1$ as the midpoint of the geodesic connection them. 

An important property of $(P_p(\Omega),W_p)$ is that it is a geodesic space and that 
geodesics are easy to characterize. Namely they are given by the \emph{displacement interpolation} (a.k.a. McCann interpolation). We first describe them for $\mu_0$ 
in $P_p(\Omega)$ which has a density: $d\mu_0 = I_0 dx$ and arbitrary $\mu_1 \in P_p(\Omega)$.
Then there exists a unique transportation map $f^*_\sharp \mu_0 = \mu_1$ which minimizes the transportation cost for the given $p \geq 1$. 
The geodesic is obtained by moving the mass at constant speed from $x$ to $f^*(x)$. 
More precisely, let $t \in [0,1]$ and $x \in \Omega$ let
\[ f^*_t(x) = (1-t) x + t f^*(x)  \]
be the position at time $t$ of the mass initially at $x$. 
Note that $f_0^*$ is identity mapping and $f_1^* = f^*$.
Pushing forward the mass by $f_t^*$: 
\[\mu^*(t) = f^*_{t \, \sharp} \mu_0 \]
provides the desired geodesic from $\mu_0$ to $\mu_1$.
We remark that the velocity of each particle $\partial_t f^*_t = f^*(x)-x$ which is nothing but the displacement of the optimal transportation map.

If $\mu_0$ does not have a density, we need to use the optimal transportation plan $\gamma \in \Gamma(\mu_0, \mu_1)$ which minimizes the $p$-transportation cost. 
For $t \in [0,1]$ let $F_t: \Omega \times \Omega \to \Omega$ be the convex interpolation: $F_t(x,y) = (1-t)x + ty$. Then $\mu^*(t) = F_{t \, \sharp} \gamma$ is the desired geodesic. The intuitive explanation is that mass from $x$ which ends up at $y$ is being transported at constant speed, along the  line segment from $x$ to $y$.
Figure \ref{fig:geodesic} conceptualizes the geodesic between two measures in $P_2(\Omega)$, and visualizes it for three different pairs of measures. 

An important fact regarding the $2$-Wasserstein space is Otto's presentation of a formal Riemannian metric for this space \cite{otto2001geometry}. It involves shifting  to Lagrangian point of view. To explain, consider path $\mu(t)$ in $P_2(\Omega)$ with smooth densities $I(x,t)$. Then $s(x,t) = \frac{\partial I}{\partial t}(x,t)$, the perturbation of $\mu(t)$, can be thought as a tangent vector. Instead of thinking of increasing/decreasing the density this perturbation can be viewed as resulting form moving the mass by a vector field.
 In other words consider vector fields $v(x,t)$ such that
\begin{equation} \label{eq:eulag}
 s = - \nabla \cdot (Iv). 
\end{equation}
There are many such vector fields. Otto defined the size of $s(\: \cdot \:,t)$ as the (square root of) the minimal kinetic energy of the vector field that produces the perturbation to density $s$.   
That is 
\begin{equation} \label{eq:rm}
 \langle s, s \rangle  = \min_{v \textrm{ satisfies } \eqref{eq:eulag}} \, \int |v|^2 I dx
\end{equation}
%
%
Utilizing the Riemmanian manifold structure of $P_2(\Omega)$ together with the inner product presented in Equation \eqref{eq:rm} the $2$-Wasserstein metric can be reformulated into finding the minimizer of the following action among all curves in $P_2(\Omega)$ connecting $\mu$ and $\nu$ \cite{benamou2000computational}, 
\begin{align}
W^2_2(\mu,\nu) =& \operatorname{inf}_{I,\text{v}} \int_{0}^{1}\int_{\Omega}I(x,t)|\text{v}(x,t)|^2 dxdt\nonumber\\
\textrm{such that } & ~ \partial_t I+ \nabla\cdot (I \text{v})=0\nonumber\\
&~  I(\cdot,0)=I_0(\cdot),~I(\cdot,1)=I_1(\cdot)
\label{eq:fluid}
\end{align}
where the first constraint is the continuity equation.

\subsection{\bf Optimal Transport: Embeddings and Transforms}

The optimal transport problem and specifically the $2$-Wasserstein metric and the Sliced $2$-Wasserstein metric have been recently used to define bijective nonlinear  transforms for signals and images  \cite{wang2013linear, park2015cumulative, kolouri2016radon,kolouri2016continuous}. In contrast to commonly used linear signal transformation frameworks (e.g. Fourier and Wavelet transforms) which only employ signal intensities at fixed coordinate points, thus adopting an `Eulerian' point of view, the idea behind the transport-based transforms is to consider the intensity variations together with the locations of the intensity variations in the signal. Therefore, such transforms adopt a `Lagrangian' point of view for analyzing signals.  Here we briefly describe these transforms and some of their prominent properties. 

\begin{figure}[t!]
\centering
\includegraphics[width=\linewidth]{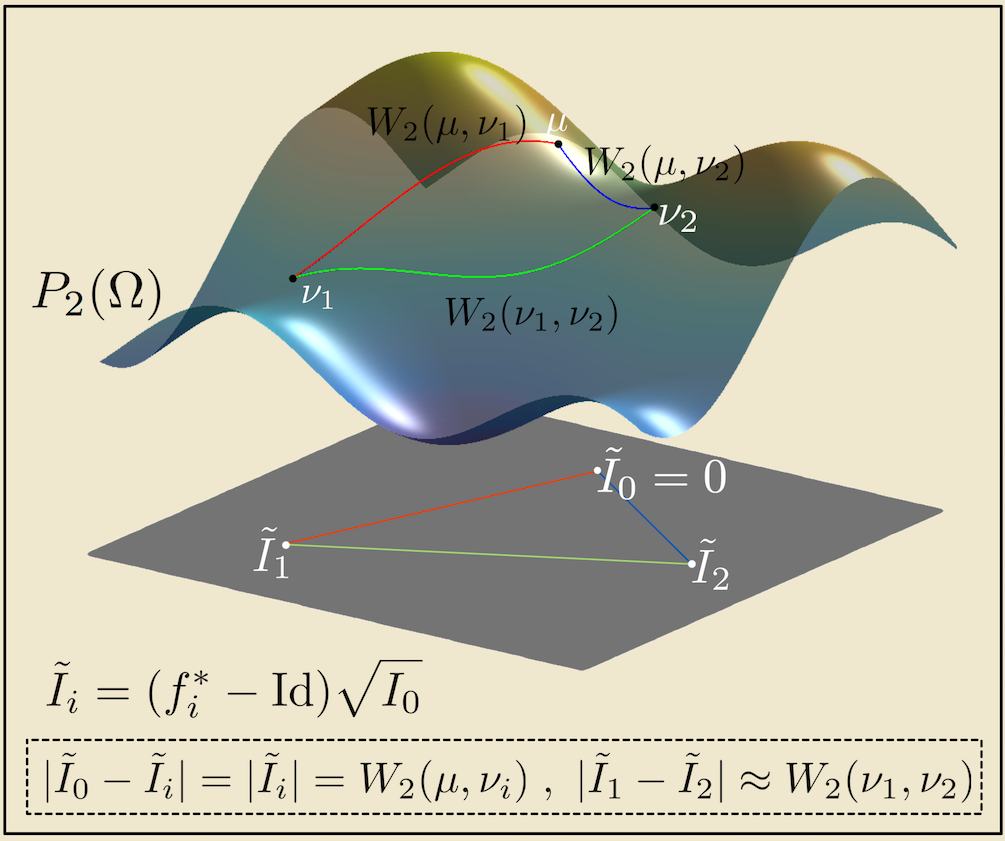}
\caption{Graphical representation of the LOT framework. The framework embeds the probability measures $\nu_i$ in the tangent space of $P_2(\Omega)$ at a fixed probability measure $\mu$ with density $I_0$. As a consequence, the Euclidean distance between the embedded functions $\tilde{I}_1$ and $\tilde{I}_2$ provides an approximation for the 2-Wasserstein distance, $W_2(\nu_1,\nu_2)$.}
\label{fig:LOT}
\end{figure}

\subsubsection{\textit{\textbf{The linear optimal transportation framework}}}
\label{sec:LOT}
The linear optimal transportation (LOT) framework was  proposed by Wang et al. \cite{wang2013linear}. The framework was successfully used in \cite{basu2014detecting,tosun2014novel, ozolek2014accurate, tosun2015detection} for pattern recognition in biomedical images and specifically histopathology and cytology images. Later, it was extended in \cite{kolouri2016continuous} as a generic framework for pattern recognition and was used in \cite{kolouri2015transport} for single-frame super-resolution reconstruction of face images. The framework was modified by Seguy and Cuturi \cite{seguy2015principal}, where they studied the principal geodesics of probability measure. The LOT framework provides an invertible Lagrangian transform for images. It was initially proposed as a method to simultaneously amend the computationally expensive requirement of calculating pairwise 2-Wasserstein distances between $N$ probability measures for pattern recognition purposes, and to allow for the construction of generative models for images involving textures and shapes.  For a given set of probability measures $\nu_i \in P_2(\Omega)$, for $i=1,...,N$, and a fixed template probability measure $\mu$, the transform projects the probability measures to the tangent space at $\mu$, $T_\mu$. The projections are acquired by finding the optimal velocity fields corresponding to the optimal transport plans between $\mu$ and each probability measure in the set.

 Formally, the framework  provides a linear embedding for $P_2(\Omega)$ with respect to a fixed measure $\mu\in P_2(\Omega)$. Meaning that, the Euclidean distance between the embedded measures, $\tilde{\nu}_i$, and the fixed measure, $\mu$, is equal to $W_2(\mu,\nu_i)$ and the Euclidean distance between two embedded measures is, generally speaking, an approximation of their 2-Wasserstein distance. The geometric interpretation of the LOT framework is presented in Figure \ref{fig:LOT}. The linear embedding then facilitates the application of linear techniques such as principal component analysis (PCA) and linear discriminant analysis (LDA) to probability measures. 

\subsubsection{\textit{\textbf{The cumulative distribution transform}}}
Park et al. \cite{park2015cumulative} considered the LOT framework for one-dimensional probability measures, and since in dimension one the transport maps are explicit, they were able to characterize the properties of the transformed densities in the tangent space. Here, we briefly review their results. Similar to the LOT framework, let $\nu_i$ for $i=1,...,N$, and $\mu$ be absolutely continuous probability measures defined on $\R$, with corresponding positive probability densities $I_i$ and $I_0$. The framework first calculates the optimal transport maps between $I_i$ and $I_0$ using $f_i(x)= F_{\nu_i}^{-1}\circ F_\mu(x)$ for all $i=1,...,N$. Then the forward and inverse transport-based transform, denoted as the cumulative distribution transform (CDT) by Park et al. \cite{park2015cumulative}, for these density functions with respect to the fixed template $\mu$ is defined as, 
\begin{eqnarray}
\left\{ \begin{array}{ll}
\tilde{I_i}=(f_i-\mathrm{Id})\sqrt{I_0} & \text{(Analysis)}\\
I_i= (f_i^{-1})' (I_0\circ f_i^{-1})& \text{(Synthesis)}
\end{array}
\right.
\end{eqnarray}
where $(I_0 \circ f^{-1}_i)(x) = I_0(f^{-1}_i(x))$. Note that the $L_2$ norm of the transformed signals, $\tilde{I}_i$ corresponds to the $2$-Wasserstein distance between $\mu$ and $\nu_i$. In contrast to the higher-dimensional LOT, the Euclidean distance between two transformed (embedded) signals $\tilde{I}_i$ and $\tilde{I}_j$, however, is the exact $2$-Wasserstein distance between $\nu_i$ and $\nu_j$ (See \cite{park2015cumulative} for a proof) and not just an approximation. Hence, the transformation is isometric with respect to the $2$-Wasserstein metric. This isometric nature of the CDT was utilized in \cite{kolouri2015sliced} to provide positive definite kernels for n-dimensional measures. 

\begin{figure*}[t!]
\centering
\includegraphics[width=.95\linewidth]{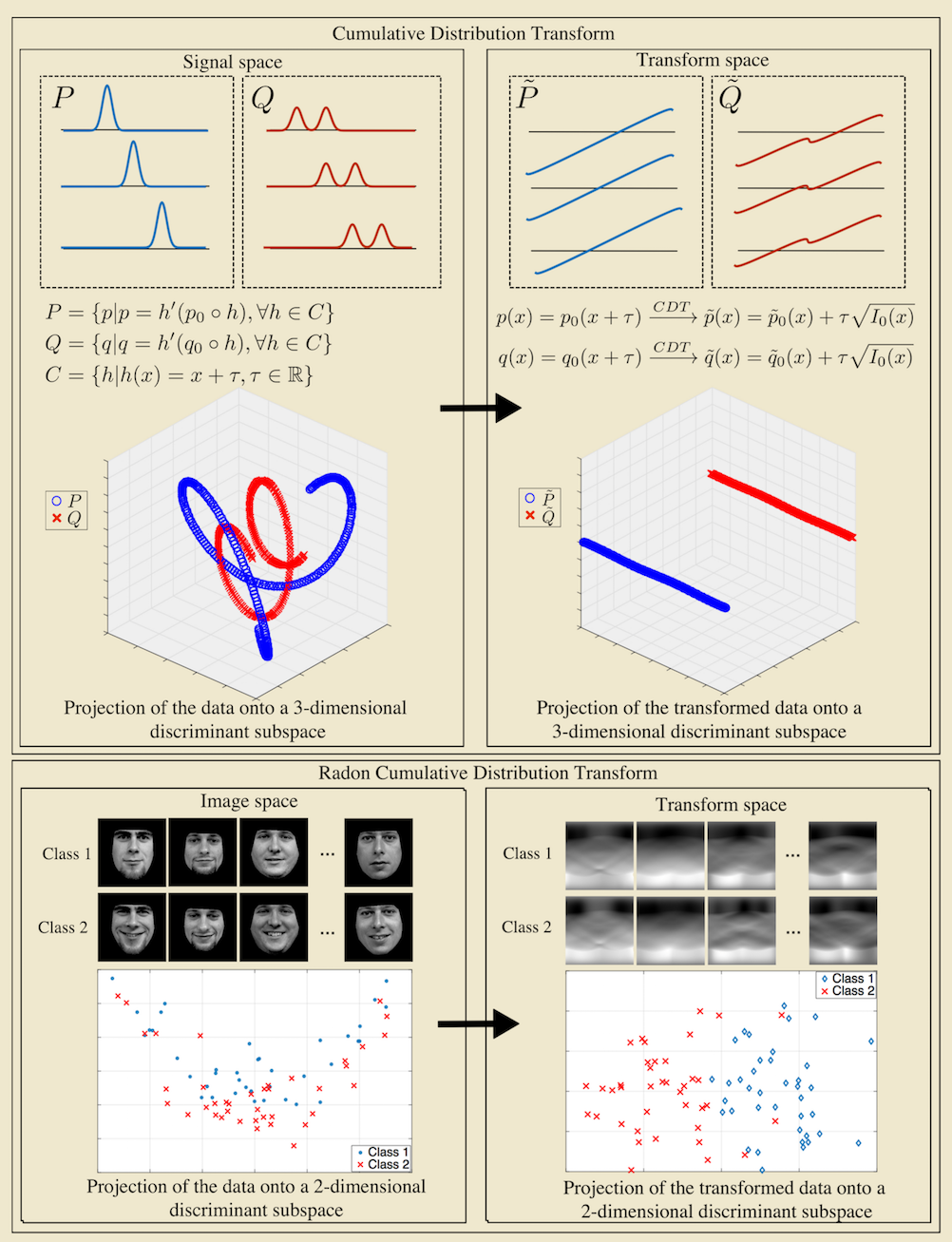}
\caption{Examples for the linear separability characteristic of the CDT and the Radon-CDT. The discriminant subspace for each case is calculated using the penalized-linear discriminant analysis ({\it p}-LDA). It can be seen that the nonlinear structure of the data is well captured in the transform spaces.}
\label{fig:CDT}
\end{figure*}

From a signal processing point of view,  the CDT is a nonlinear signal transformation that captures certain nonlinear variations in signals including translation and scaling. Specifically, it gives rise to the  transformation pairs presented in Table \ref{tab:CDT}.
\begin{table}[t!]
\centering
{\rowcolors{3}{background}{backgroundd}
\begin{tabular}{ |p{.22\columnwidth}|p{.25\columnwidth}|p{.45\columnwidth}|  }
\hline
\multicolumn{3}{|c|}{Cumulative Distribution Transform pairs} \\
\hline
Property & Signal domain & CDT domain\\
\hline
 &   $I(x)$ &$\tilde{I}(x)$ \\
Translation & $I(x-\tau)$   & $\tilde{I}(x)+\tau\sqrt{I_0(x)}$ \\
Scaling &$  a I(a x)$ & $\frac{\tilde{I}(x)}{a} - x\frac{(a-1)}{a}\sqrt{I_0(x)}$ \\
Composition    &$g'(x)I(g(x))$ & $(g^{-1}(\frac{\tilde{I}(x)}{\sqrt{I_0(x)}}+x)-x)\sqrt{I_0(x)}$ \\
\hline
\end{tabular}
}
\caption{Cumulative Distribution Transform pairs. Note that the composition holds for all strictly monotonically increasing functions $g$.}
\label{tab:CDT}
\end{table}
%
From Table \ref{tab:CDT} one can observe that although $I(t-\tau)$ is nonlinear in $\tau$, its CDT representation $\tilde{I}(t)+\tau\sqrt{I_0(t)}$ becomes affine in $\tau$ (similar effect is observed for scaling). In effect, the Lagrangian transformations (compositions) in original signal space are rendered into Eulerian perturbations in transform space, borrowing from the PDE parlance. Furthermore, Park et al. \cite{park2015cumulative} demonstrated that the CDT facilitates certain pattern recognition problems. More precisely, the transformation turns certain not linearly separable and disjoint classes of signals into linearly separable ones. Formally, let $C$ be a set of measurable maps and let $P,Q\subset P_2(\Omega)$ be sets of positive probability density functions born from two positive probability density functions $p_0,q_0\in P_2(\Omega)$ (mother density functions) as follows,
\begin{eqnarray}
P&=&\{p| p=h'(p_0\circ h), \forall h\in C \},\nonumber\\
Q&=&\{q| q=h'(q_0\circ h), \forall h\in C \}.
\end{eqnarray}
The sets $P$ and $Q$ are disjoint but not necessarily linearly separable in the signal space. A main result of \cite{park2015cumulative} states that the signal classes $P$ and $Q$ are guaranteed to be linearly separable in the transform space (regardless of the choice of the reference signal $I_0$) if $C$ satisfies the following conditions, 
\begin{enumerate}[i)]
\item $h\in C \iff h^{-1}\in C$
\item $h_1, h_2\in C \Rightarrow \rho h_1+(1-\rho) h_2 \in C, ~\forall\rho\in[0,1]$
\item $h_1, h_2\in C \Rightarrow h_1(h_2), h_2(h_1)\in C$ 
\item $h'(p_0\circ h)\neq q_0,~\forall h\in C$ 
\end{enumerate}
The set of translations $C=\{f|f(x)=x+\tau, \tau\in \R\}$, and scaling $C=\{f|f(x)=ax, a\in \R^+\}$, for instance, satisfy above conditions. We refer the reader to \cite{park2015cumulative} for further reading. The top panel in Figure  \ref{fig:CDT} demonstrates the linear separation property of the CDT and demonstrate the capability of such nonlinear transformation. The signal classes $P$ and $Q$ are chosen to be the set of all translations of a single Gaussian and a Gaussian mixture including two Gaussian functions with a fixed mean difference, respectively. The discriminant subspace is calculated for these classes and it is shown that while the signal classes are not linearly separable in the signal domain, they become linearly separable in the transform domain.

\subsubsection{\textit{\textbf{The Radon cumulative distribution transform}}}
\begin{figure}[t!]
\centering
\includegraphics[width=\linewidth]{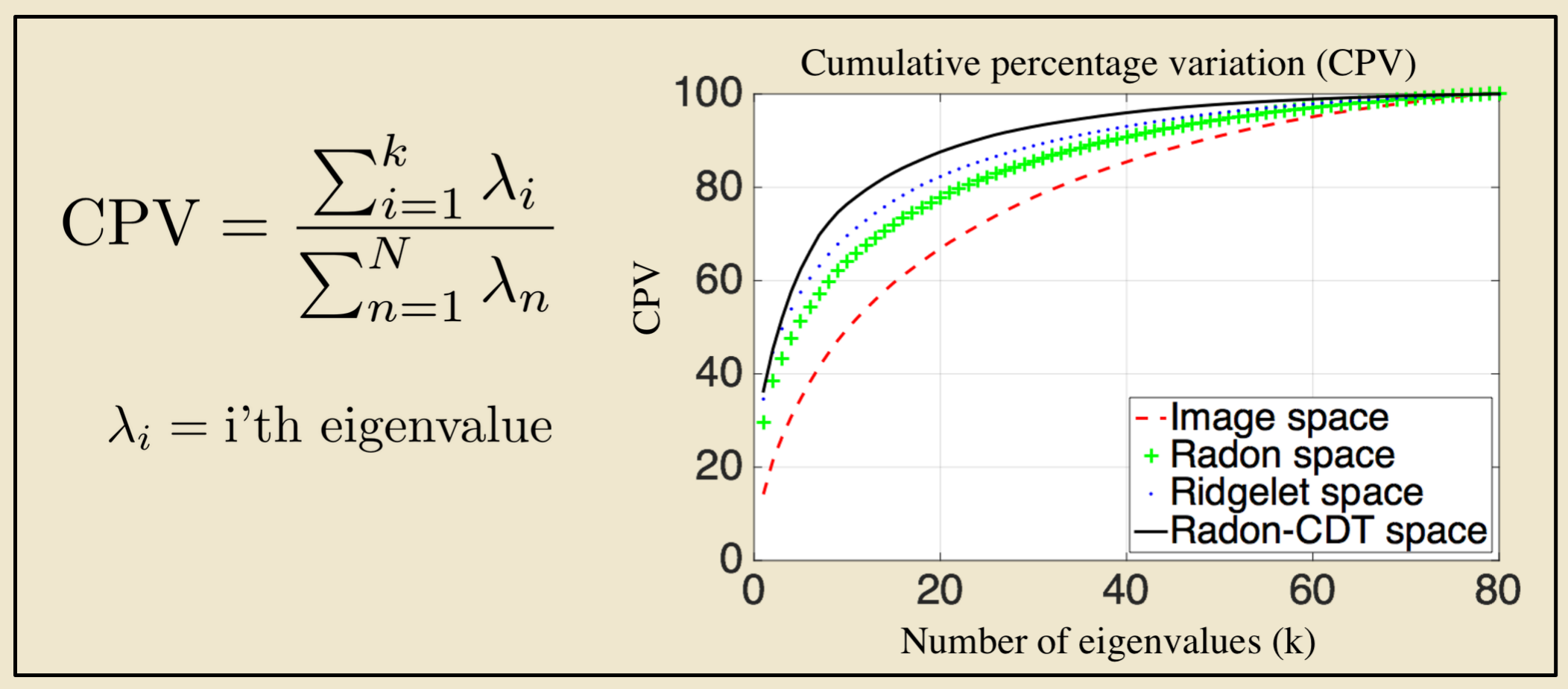}
\caption{The cumulative percentage of the face dataset in Figure \ref{fig:CDT} in the image space, the Radon transform space, the Ridgelet transform space, and the Radon-CDT transform space.}
\label{fig:CPV}
\end{figure}

The CDT framework was extended to higher dimensional density functions through the sliced-Wasserstein distance in  \cite{kolouri2016radon}, and was denoted as the Radon-CDT. It is shown in \cite{kolouri2016radon} that similar characteristics of the CDT, including the linear separation property, also hold for the Radon-CDT. Figure  \ref{fig:CDT}  clarifies the linear separation property of the Radon-CDT and demonstrate the capability of such transformations.  Particularly, Figure \ref{fig:CDT} shows a facial expression dataset with two classes (i.e. neutral and smiling expressions) and its corresponding representations in the LDA discriminant subspace calculated from the images (bottom left panel), the Radon-CDT of the dataset and the corresponding representation of the transformed data in the LDA discriminant subspace (bottom right panel). It is clear that the image classes become more linearly separable in the transform space. In addition, the cumulative percentage variation of the dataset in the image space, the Radon transform space, the Ridgelet transform space, and the Radon-CDT space are shown in Figure \ref{fig:CPV}. It can be seen that the variations in the dataset could be explained with fewer components in the nonlinear transform spaces as opposed to the linear ones.

\section{Numerical methods}
\label{sec:numerics}

There exists a variety of fundamentally different approaches to finding optimal transportation maps and plans. Below we present the notable approaches.
In the table \ref{tab:numer} we summarize the expected computational complexity of the given algorithm. 

\begin{table}[t!]
\centering
{\rowcolors{3}{background}{backgroundd}
\begin{tabular}{ |p{.32\columnwidth}|p{.10\columnwidth}|p{.54\columnwidth}|  }
\hline
\multicolumn{3}{|c|}{Comparison of Numerical Approaches} \\
\hline
Method & Compl. & Remark \\
linear programming & $N^3$ & Applicable to general costs. Good approach if the measures are supported at very few sites \\
multi-scale linear programming & $N$ &  
Applicable to general costs. Fast and robust method, though truncation involved can lead to imprecise distances.  \\
Auction algorithm & $N^2$  & Applicable only when the number of particles in the source and the target is equal and all of their masses are the same. \\
 Entropy regularized linear programing & $N$ & Applicable to general costs. Simple and performs very well in practice for moderately large problems. Difficult to obtain high accuracy. \\
Fluid mechanics &  $N^{1+1/d}$ & This approach can be adapted to generalizations of the quadratic cost, based on action along paths. \\
AHT  minimization & $N$ 
& Quadratic cost. Requires some smoothness and positivity of densities. Convergence is only guaranteed for infinitesimal stepsize. \\
Gradient descent on the dual problem & $N$ & Quadratic cost. Convergence depends on the smoothness of the densities, hence a multi-scale approach is needed for non-smooth densities (i.e. normalized images).\\
Monge--Ampere solver & $N$ &  Quadratic cost. One in \cite{benamou2014numerical} is proved to be convergent. Accuracy is an issue due to wide stencil used.  \\
 \hline
\end{tabular}}
\caption{
The numerical complexity and key properties of various numerical approaches. Each measure consists of about  $N$ delta masses in $\R^d$.
(So if we are dealing with an image this would mean that $N$ is the total number of pixels and $d=2$). Complexities are reported with $\log N$ factors neglected. 
\color{black} }
\label{tab:numer}
\end{table}

\subsection{\bf Linear programming}
Leonid Kantorovich's extensive monologue on `Mathematical Methods in the Organization and Planning of Production' in 1939 \cite{kantorovich1960mathematical}, George B Dantzig's pioneering work on the simplex algorithm in 1947 \cite{dantzig1948programming}, John von Neumann's duality theory in 1947 \cite{von1947theory}, and Tjalling Koopmans' work on `Activity analysis of production and allocation' in 1951 \cite{koopmans1951activity} were the founding pillars of the linear programming as we know it today. The term `linear programming', was coined by Koopmans in \cite{koopmans1951activity}. The linear programming problem, is an optimization problem with a linear objective function and linear equality and inequality constraints. The set of feasible points to a linear programming problem forms a possibly unbounded convex set. The optimal solution can be thought of as the intersection of the hyper plane corresponding to the linear objective function with an extreme point (i.e. a corner point) of the feasible convex set of solutions. 

Several numerical methods exist for solving linear programming problems \cite{vanderbei2014linear}, among which are the simplex method \cite{dantzig1948programming} and its variations \cite{cunningham1976network,dantzig2006linear,bazaraa2011linear} and the interior-point methods \cite{karmarkar1984new,lustig1994interior}. The computational complexity of the mentioned numerical methods, however, scales at best cubically in the size of the domain. Hence, assuming the density measures have $N$ particles the number of unknowns $\gamma_{ij}$s is $N^2$ and the computational complexities of the solvers are at best $\O(N^3 log N)$ \cite{rubner2000earth,cuturi2013sinkhorn}. The computational complexity of the linear programming methods is a very important limiting factor for the applications of the Kantorovich problem.

We point out that in the special case where $M=N$ and the mass is equidistributed, the optimal transport problem simplifies to a one to one assignment problem, which can be solved by the auction algorithm of Bertsekas \cite{bertsekas1988auction} or similar methods in $\O(N^2log N)$. In addition, Pele and Werman showed that a thresholded version of the Kantorovich problem can be solved through the min-cost-flow algorithm in $\O(N^2)$ \cite{pele2009fast}. Several multiscale approaches and sparse approximation approaches have recently been introduced to improve the computational performance of the linear programming solvers, including Schmitzer \cite{schmitzer2015sparse}, ]. The work by Oberman and Ruan \cite{oberman2015efficient} provides an algorithm which is claimed to be of linear complexity for the linear programming problem. Here we provide a brief description of this method.
\subsubsection{\textit{\textbf{Multi-Scale algorithm}}}
For a coarse grid of size $h$, the space $\Omega\subset\mathbb{R}^d$ is uniformly discretized into hypercubes with centers $x_i$ presented with the set $\Omega_h=\{x_i\}_{i=1}^N$ and where the distance between neighboring points is $h$. Measures $\mu$ and
$\nu$ are discretized by $\mu^h = \sum_{i=1}^N p_i \delta_{x_i}$ and $\nu^h = \sum_{j=1}^Nq_i \delta_{x_i}$. One then performs the linear program~\eqref{eq:lp} with inputs $p_i = p_i^h$, $q_i = q_i^h$, $x_i = x_i^h$ and $y_i = y_i^h$ to calculate the optimal map $\gamma^{h}$ between $\mu^h$ and $\nu^h$.

In the next step, the discretization is refined to $\frac{h}{2}$. Since  $\gamma^h$ is expected to converge to  the optimal transport plan between  $\mu$ and $\nu$,  $\gamma^*$, then $\gamma^{\frac{h}{2}}$ should be close to $\gamma^h$. Let $S_h$ be the support of $\gamma^h$ on the grid $\Omega_h\times \Omega_h$ and $\bar{S}_h$ be the set of neighbors (in space) of $S_h$. The Oberman and Ruan method assumes that 
\[ S_{\frac{h}{2}} \subseteq P_{h\mapsto\frac{h}{2}} (\bar{S}_h) \]
where $P_{h\mapsto \frac{h}{2}}$ is the projection onto the refined grid $\Omega_{\frac{h}{2}}^2$ (where one point in $\Omega_h$ is projected to $2^d$ points in $\Omega_{\frac{h}{2}}$). One then solves the linear program on the refined grid. For better resolution repeat the refinement procedure. 
The reason that the multi-scale procedure is advantageous is that the linear programming problem 
solved at the finest level is expected to have $O(N)$ variables as opposed to $O(N^2)$ variables 
that the original full linear program has. 

%

 \begin{figure}[t!]
\centering
\includegraphics[width=\linewidth]{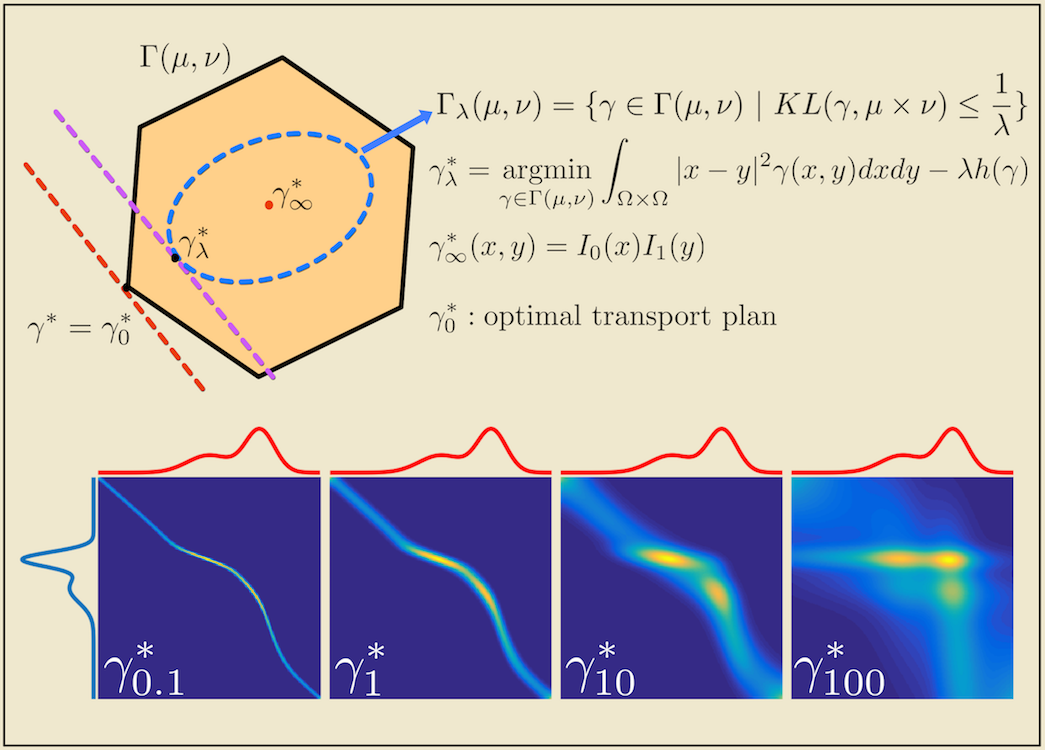}
\caption{The geometric interpretation of the entropy regularization of the Kantorovich problem. For $\lambda=0$ the optimal transport plan is located at an extreme point (corner point) of the feasible set $\Gamma(\mu,\nu)$. Increasing $\lambda$ corresponds to constraining the space to the convex subset of $\Gamma(\mu,\nu)$, for which the KL-divergence between the transport plans and  $\gamma^*_\infty=\mu\times\nu$ is smaller than $\frac{1}{\lambda}$. The entropy regularized optimal transport plan, $\gamma^*_\lambda$ is shown for a one-dimensional example with different values of $\lambda$. }
\label{fig:KL}
\end{figure}

\subsection{\bf Entropy regularized solution}
Cuturi's  work \cite{cuturi2013sinkhorn} provides a fast and easy to implement variation of the Kantorovich problem, which has attracted ample attention recently.  Cuturi \cite{cuturi2013sinkhorn} proposed a variation of the Kantorovich problem by considering the transportation problem from a maximum-entropy perspective. He suggested to regularize the Wasserstein metric by the entropy of the transport plan. This modification simplifies the problem and enables much faster numerical schemes
{with complexity $\O(N^2)$ \cite{cuturi2013sinkhorn} or $\O(Nlog(N))$ using the convolutional Wasserstein distance presented in \cite{solomon2015convolutional}
 (compared to  $\O(N^3)$ of the linear programming methods), 
where $N$ is the number of delta masses in each of the measures. 
The price one pays is that it is difficult to obtain high accuracy approximations 
 of the optimal transport plan.}  Formally, the entropy regularized p-Wasserstein distance, otherwise known as the Sinkhorn distance, between probability measures $\mu$ and $\nu$ defined on the metric space $(\Omega,d)$ is defined as, 
\begin{eqnarray}
W_{p,\lambda}^p(\mu,\nu)=\operatorname*{inf}_{\gamma\in\Gamma(\mu,\nu)}  \int_{\Omega\times \Omega} d^p(x,y)\gamma(x,y)dxdy-\lambda h(\gamma) 
\label{eq:entropyreg}
\end{eqnarray}
where $h(\gamma)$ is the entropy of the plan and is defined as, 
\begin{eqnarray}
h(\gamma)=-\int_{\Omega\times\Omega}\gamma(x,y) \ln(\gamma(x,y))dxdy.
\end{eqnarray}
We note that this is not a true metric since $W_{p,\lambda}^p(\mu,\mu)>0$. 
Since the entropy term is strictly concave, the overall optimization problem in \eqref{eq:entropyreg} becomes strictly convex. It is straightforward to show that ( see \cite{cuturi2013sinkhorn})  the entropy regularized p-Wasserstein distance in Equation \eqref{eq:entropyreg} can be reformulated as,
\begin{eqnarray}
W_{p,\lambda}^p(\mu,\nu)=\lambda \operatorname*{inf}_{\gamma\in\Gamma(\mu,\nu)} \text{KL}(\gamma|\K_\lambda)
\end{eqnarray}
where $\K_\lambda(x,y)=\exp(-\frac{d^p(x,y)}{\lambda})$ and $\text{KL}(\gamma|\K_\lambda)$ is the Kullback-Leibler (KL) divergence between $\gamma$ and $\K_\lambda$ and is defined as, 
\begin{eqnarray}
\text{KL}(\gamma|\K_\lambda)=\int_{\Omega\times\Omega} \gamma(x,y) \ln(\frac{\gamma(x,y)}{\K_\lambda(x,y)})dxdy.
\label{eq:sinkhorn}
\end{eqnarray}
Figure \ref{fig:KL} provides a geometric interpretation of the entropy regularization as discussed in \cite{cuturi2013sinkhorn}. It can be seen that, the regularizer enforces the plan to be within $\frac{1}{\lambda}$ radius in the KL-divergence sense from the transport plan $\gamma^*_\infty=\mu\times \nu$. 

In the discrete setting, where  $\mu=\sum_{i=1}^N p_i\delta_{x_i}$ and $\nu=\sum_{j=1}^N q_j \delta_{y_j}$,  $\gamma$ and $\K_\lambda$ are $N\times N$ matrices and hence the number of unknowns is $N^2$. However, using Equation \eqref{eq:sinkhorn} it can be shown that the optimal transport plan $\gamma$ is of the form $D_v\K_\lambda D_w$ where $D_v$ and $D_w$ are diagonal matrices with diagonal entries $v,w\in \R^N$ \cite{cuturi2013sinkhorn}. Therefore, due to the new formulation the number of unknowns decreases from $N^2$ to $2N$. The new problem can then be solved through the iterative proportional fitting procedure (IPFP) \cite{deming1940least,ruschendorf1995convergence}, otherwise known as the RAS algorithm \cite{bacharach1965estimating},  or alternatively through the Sinkhorn-Knopp algorithm \cite{sinkhorn1967concerning,knight2008sinkhorn}. The Sinkhorn-Knopp algorithm is specifically interesting as it is computationally efficient and very simple to implement.

The entropy regularization of the transportation problem has recently attracted ample attention from both application and numerical analysis point of view. One can refer to Benamou et al. \cite{benamou2015iterative} and Solomon et al. \cite{solomon2015convolutional} for further numerical analysis, and Cuturi and Doucet \cite{cuturi2013fast}, Rabin et al. \cite{rabin2015convex}, and Rabin and Papadakis \cite{rabin2014non} and others for applications of the entropy regularized transportation problem. It should be mentioned, however, that the number of iterations for the entropy regularized solvers increases as the regularization parameter goes to zero \cite{oberman2015efficient}. In addition, these method become numerically unstable for small regularization parameters \cite{benamou2015iterative, oberman2015efficient}. This is due to the fact that the elements of matrix $\K_\lambda$ go to zero as $\lambda$ goes to zero.

\subsection{\bf Fluid dynamics solution}

Benamou and Brenier \cite{benamou2000computational} presented a fluid dynamic formulation of the $2$-Wassestein metric using the continuity equation as discussed in Section  \ref{subsec:fluid} and Equation \eqref{eq:fluid}. They provided a numerical scheme for optimizing the problem in \eqref{eq:fluid}, using the augmented Lagrangian method \cite{fortin2000augmented}. A brief overview of the method and its derivation is as follows.  Let $\phi(x,t)$ be the space-time dependent Lagrange multiplier for constraints in Equation \eqref{eq:fluid}. Using $\phi(x,t)$ the Lagrangian is given by, 
\begin{eqnarray}
L(\phi,\rho,\text{v})&=& \int_0^1\int_{\R^d} I |\text{v}|^2+\phi(\partial_tI+\nabla\cdot(I\text{v}))dxdt\nonumber\\
&=&\int_0^1\int_{\R^d} \frac{|m|^2}{2I}-\partial_t\phi I-\nabla_x\phi\cdot m dxdt\nonumber\\
&&~-\int_{\R^d} \phi(0,\cdot)I_0- \phi(1,\cdot)I_1dx
\label{eq:lagrangian}
\end{eqnarray}
where $m=I \text{v}$ and we used integration by parts together with the equality constraints in Equation \eqref{eq:fluid} to obtain the second line. Note that, 
\begin{eqnarray}
W^2_2(\mu,\nu)=\operatorname*{inf}_{\rho,m}\operatorname*{sup}_{\phi} L(\phi,I,\text{v}).
\label{eq:lagrangianOT}
\end{eqnarray}
Using the Legendre transform \cite{boyd2004convex} on $|m|^2/2I$ one can write, 
\begin{eqnarray}
\frac{|m(x,t)|^2}{2I(x,t)}&=& \operatorname*{sup}_{a,b}~~ a(x,t)I(x,t)+b(x,t)m(x,t)\nonumber\\
&& s.t.~~~ a(x,t)+\frac{|b(x,t)|^2}{2} \leq 0, ~ \forall x,t
\label{eq:legendre}
\end{eqnarray}
Define $\psi:=\{I,m\}$, $q:=\{a,b\}$, and their corresponding inner product to be,
\begin{equation*}
\langle \psi , q\rangle=\int_0^1\int_{\R^d}  I(x,t) a(x,t)+ m(x,t)\cdot b(x,t)~ dx dt,
\end{equation*}
Also define ,
\begin{eqnarray*}
F(q)&:=&\left\{ \begin{array}{lc}
0 & a(x,t)+\frac{|b(x,t)|^2}{2} \leq 0, ~ \forall x,t\\
+\infty & O.W.
\end{array} \right. \\
G(\phi)&:=&\int_{\R^d} \phi(0,\cdot)I_0- \phi(1,\cdot)I_1dx
\end{eqnarray*}
then it is straightforward to show that Equation \eqref{eq:lagrangianOT} can be rewritten as, 
\begin{eqnarray}
W^2_2(\mu,\nu)=\operatorname*{sup}_{\psi}\operatorname*{inf}_{\phi,q} ~F(q)+G(\phi)+\langle \psi, \nabla_{t,x}\phi-q\rangle
\end{eqnarray}
where $\nabla_{t,x}\phi=\{ \partial_t \phi,\nabla_{x}\phi\}$. In the formulation above, $\psi$ can be treated as the Lagrange multiplier for a new constraint, namely $\nabla_{t,x}\phi=q$. Thus the augmented Lagrangian can be written as, 
\begin{eqnarray}
L_r(\phi,q,\psi)&=& F(q)+G(\phi)+\langle \psi, \nabla_{t,x}\phi-q\rangle +\nonumber \\ 
&&~ \frac{r}{2} \langle \nabla_{t,x}\phi-q, \nabla_{t,x}\phi-q\rangle
\end{eqnarray}
and finally the corresponding saddle point problem is, 
\begin{eqnarray}
W^2_2= \operatorname*{sup}_{\psi}\operatorname*{inf}_{\phi,q} L_r(\phi,q,\psi)
\label{eq:saddle}
\end{eqnarray}
Benamou and Brenier \cite{benamou2000computational} used a variation of the Uzawa algorithm \cite{elman1994inexact} to solve the problem above.  We note that recent methods based on Bregman iteration such as \cite{zhang2011unified,esser2009applications} could also be used for solving the saddle point problem in \eqref{eq:saddle}. 

Due to the space-time nature of the fluid dynamic formulation, the solvers require a space-time discretization (spatial-temporal grid), which increases the computational complexity of such solvers. However, the fluid dynamics solution enables one to handle situations for which there exist barriers in the domain. Take for instance transportation of a distribution through a maze, in which the mass can not be transported over the maze walls. 

\subsection{\bf Flow minimization (AHT)}

Angenent, Haker, and Tannenbaum (AHT) \cite{angenent2003minimizing} proposed a flow minimization scheme to obtain the optimal transport map from the Monge problem. The method was used in several image registration applications \cite{haker2004optimal,zhu2007image}, pattern recognition \cite{kolouri2016continuous,wang2013linear}, and computer vision \cite{kolouri2015transport}. The method was polished in several follow up papers \cite{haker2004optimal,ur20093d}. The idea behind the method, is to first obtain an initial mass preserving transport map using the Knothe-Rosenblatt coupling \cite{rosenblatt1952remarks, villani2008optimal} and then update the initial map to obtain a curl free mass preserving transport map that minimizes the transport cost. A brief review of the method is provided here. 

Let $\mu$ and $\nu$ be continuous probability measures defined on convex domains $X,Y\subseteq \R^d$ with corresponding positive densities $I_0$ and $I_1$. In order to find the optimal transport map, $f^*$, AHT starts with an initial transport map, $f_0:X\rightarrow Y$ calculated from the Knothe-Rosenblatt coupling \cite{rosenblatt1952remarks, villani2008optimal}, for which $(f_0)_\#\mu=\nu$. Then it updates $f_0$ to minimize the transport cost. The goal, however, is to update $f_0$ in a way that it remains a transport map from $\mu$ to $\nu$. 
AHT defines $s(x,t)$, where for a fixed time, $t_0$, $s(x,t_0):X\rightarrow X$ is a transport map from $\mu$ to itself. The initial transport map is then updated through $s(x,t)$, starting from $s(x,0)=x$, such that $f_0(s^{-1}(x,t))$ minimizes the transport cost. Following simple calculations, one can show (see \cite{haker2004optimal}) that for $s(x,t)$ to be a MP mapping from $I_0$ to itself, $\frac{\partial s}{\partial t}$ should have the following form,
\begin{eqnarray}
\frac{\partial s}{\partial t}=(\frac{1}{I_0} \xi ) \circ s,
\label{eq:supdate}
\end{eqnarray}
for some vector field $\xi$ on $X$ with $div(\xi)=0$ and $\langle \xi,n\rangle=0$ on the boundary, where $n$ is the vector normal to the boundary. From (\ref{eq:supdate}), it follows that the time derivative of $f(x,t)=f_0(s^{-1}(x,t))$ satisfies,
\begin{eqnarray}
\frac{\partial f}{\partial t}=-\frac{1}{I_0} (D f) \xi.
\label{eq:fupdate}
\end{eqnarray}
AHT then differentiate the Monge objective function, 
\begin{eqnarray}
M(f)=\int_X |f(x,t)-x|^2I_0(x) dx,
\end{eqnarray}
with respect to $t$, which after rearranging will lead to,
\begin{eqnarray}
\frac{\partial M}{\partial t}=-2\int_X \langle f(x,t),I_0(x)(\frac{\partial s}{\partial t} \circ s^{-1}(x,t)) \rangle dx.
\end{eqnarray}
Substituting (\ref{eq:supdate}) in the above equation we get,
\begin{eqnarray}
\frac{\partial M}{\partial t}=-2\int_\Omega \langle f(x,t),\xi(x,t) \rangle dx.
\end{eqnarray}
Writing the Helmholtz decomposition of $f$ as $f=\nabla \phi + \chi$ and using the divergence theorem we get,
\begin{eqnarray}
\frac{\partial M}{\partial t}=-2\int_\Omega \langle \chi(x,t),\xi(x,t) \rangle dx,
\label{eq:Mt}
\end{eqnarray}
thus, $\xi=\chi$ decreases $M$ the most. In order to find $\chi$, AHT first find $\phi$ and then subtract its gradient from $f$. Given that $div(\chi)=0$ and $\langle \chi,n \rangle=0$  and taking the divergence of the Helmholtz decomposition of $f$ leads to,
\begin{eqnarray}
\left\{
\begin{array}{l}
\Delta(\phi)=div(f)\\
\langle \nabla \phi , n\rangle=\langle f , n\rangle \:\: \textrm{on the boundary} 
\end{array}
\right.
\label{eq:phi}
\end{eqnarray}
where $\Delta$ is the Laplace operator. Therefore, $\xi=f-\nabla(\Delta^{-1} div(f))$ where $\Delta^{-1} div(f)$ is the solution to (\ref{eq:phi}). Substituting $\xi$ back into (\ref{eq:fupdate}) we have,
\begin{eqnarray}
\frac{\partial f}{\partial t}=-\frac{1}{I_0} Df (f-\nabla(\Delta^{-1} div(f))),
\label{eq:fupdate1}
\end{eqnarray}
which for $f\in \R^2$ can be even simplified further,
\begin{eqnarray}
\frac{\partial f}{\partial t}=-\frac{1}{I_0} Df \nabla^\perp\Delta^{-1} div(f^\perp).
\label{eq:fupdate2}
\end{eqnarray}
where $\perp$ indicates rotation by $90$ degrees. Finally, the transport map $f$ is updated with a gradient descent scheme, 
\begin{eqnarray}
f(x,t+1)=f(x,t)+\epsilon \frac{1}{I_0} Df (f-\nabla(\Delta^{-1} div(f)))
\end{eqnarray}
where $\epsilon$ is the gradient descent step size. AHT show that for infinitesimal step size, $\epsilon$, $f(x,t)$ converges to the optimal transport map. 

\subsection{\bf Gradient descent on the dual problem}

\begin{figure}
\centering
\includegraphics[width=\columnwidth]{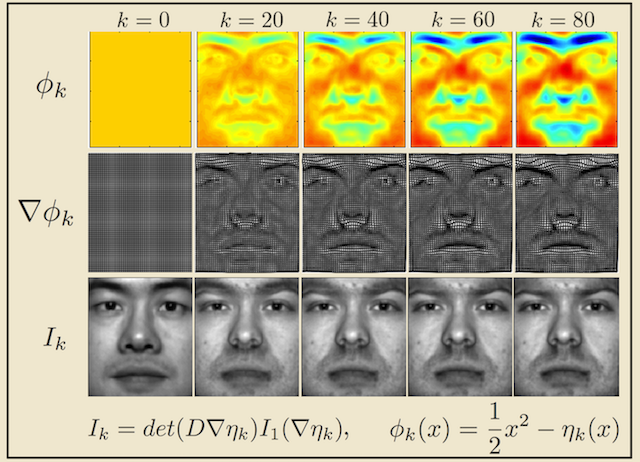}
\caption{Visualization of the iterative update of the transport potential and correspondingly the transport displacement map through CWVP iterations.}
\label{fig:chrtrnd}
\end{figure}

Chartrand, Wohlberg, Vixie, and Bollt (CWVB) \cite{chartrand2009gradient} provide a straight-forward algorithm for computing the optimal transport. It should be mentioned, however, that the method has not been studied in depth. The method reformulates the Kantorovich's dual formulation as described in Eq. \ref{eq:kdual} into minimization of a convex, continuous functional. More importantly, the derivative of the continuous functional can be explicitly found, and thus a gradient descent approach is developed for computing the optimal transport. A brief review of the method and its derivation is presented here. 

For the strictly convex cost function, $c(x,y)=\frac{1}{2}|x-y|^2$, the dual problem can be written as, 
\begin{eqnarray}
KD(\mu,\nu)= \operatorname{min}_{\eta} \underbrace{\int_X \eta(x)d\mu(x)+\int_Y \eta^c(y)d\nu(y)}_{M(\eta)}
\end{eqnarray}
where $\eta^c(y):=\operatorname{max}_{x\in X} (x\cdot y-\eta(x))$ is the Legendre-Fenchel transform of $\eta(x)$.  Note  the relationship between $\eta$ and $\phi$ in Section \ref{sec:dual} is $\phi(x)=\frac{1}{2}x^2-\eta(x)$.

The method utilizes the following property of the Legendre-Fenchel transform \cite{boyd2004convex}, 
\begin{eqnarray}
\nabla \eta^c(y) = \operatorname{argmax}_{x\in X} (x\cdot y - \eta(x))
\end{eqnarray}
and derives the functional derivative of $M$ as, 
\begin{eqnarray}
\frac{d M(\eta)}{d \eta}= I_0-\det(I-H\eta^{cc})I_1(id-\nabla \eta^{cc})
\end{eqnarray}
where $H\eta^{cc}$ is the Hessian matrix of $\eta^{cc}$, $I$ is the identity matrix, and $\eta^{cc}(x)=\operatorname{max}_{y\in Y} (y\cdot x -\eta^c(y))$. Note that if $\eta$ is a concave  function (respectively if $\phi$ is a convex function) then $\eta^{cc}=\eta$ (and $\phi^{cc}=\phi$). CWVB initializes $\eta_0(x)=\frac{1}{2}|x|^2$ and updates the potential field $\eta$ through gradient descent, 
\begin{eqnarray}
\eta_{k+1}=\eta_{k}-\epsilon_k \frac{d M(\eta)}{d \eta}
\end{eqnarray}
where $\epsilon_k$ is the step size and is calculated with a line search method. We note that when $\mu$ and $\nu$ are absolutely continuous probability measures  the optimal transport map is obtained from $\eta^*$ through, $f^*(x)=\nabla\eta^*(x)$. Figure \ref{fig:chrtrnd} visualizes the iterations of the CWVB method for two face images taken from YaleB face database (normalized to integrate to 1) \cite{croppedYaleB}. We note that the number of iterations required for convergence of the CWVB method depends on the smoothness of the input densities. In \cite{chartrand2009gradient} the authors proposed a multi-scale scheme (i.e. using Gaussian pyramids) to update $\eta$ through different levels of smoothness and hence decrease the number of iterations needed for convergence. 

\subsection{\bf Monge-Ampere equation}

The Monge-Amp\'{e}re partial differential equation (PDE) is defined as, 
\begin{eqnarray}
\det(H \phi)= h(x,\phi,D\phi)
\label{eq:mongeampere}
\end{eqnarray}
for some functional $h$, and where $H\phi$ is the Hessian matrix of $\phi$. The  Monge-Amp\'{e}re PDE is closely related to the Monge problem. According to Bernier's theorem (discussed in Section \ref{sec:brenier}) when $\mu$ and $\nu$ are absolutely continuous  probability measures on  sets $X,Y\subset\R^n$, the optimal transport map than minimizes the $2$-Wasserstein metric is uniquely characterized as the gradient of a convex function $\phi:X\rightarrow Y$. Moreover, we showed that the mass preserving constraint of the Monge problem can be written as $\det(Df)I_1(f)=I_0$. Combining these results one can write, 
\begin{eqnarray}
\det(D(\nabla \phi(x)))= \frac{I_0(x)}{I_1(\nabla \phi))}
\label{eq:otmongeampere}
\end{eqnarray}
where $D\nabla\phi=H\phi$, and therefore the equation shown above is in the form of the Monge-Amp\'{e}re PDE.  Now if $\phi$ is a convex function on $X$ satisfying $\nabla \phi(X)=Y$ and solving the Equation \eqref{eq:otmongeampere} 
then  $f^*=\nabla\phi$ is the optimal transportation map from $\mu$ to $\nu$.
 The geometrical constraint on this problem is rather unusual in PDEs and is often referred to as the optimal transport boundary conditions. 
 The equation \eqref{eq:otmongeampere} is a nonlinear elliptic equation and thus one hopes that methods achieving complexity $O(N)$ (as do the solvers for the Laplace equation) can be designed.
 Several authors have proposed numerical methods to obtain the optimal transport map through solving the Monge-Amp\'{e}re PDE in Equation \eqref{eq:otmongeampere} \cite{loeper2005numerical,trudinger2006second,caffarelli2010free,saumier2010efficient,sulman2011efficient,benamou2014numerical}. 
In particular the scheme in \cite{benamou2014numerical} is monotone, has compexity $O(N)$ (up to logarithms) and is provably convergent. 
We conclude by remarking that several regularity results on the optimal transport maps, including the  pioneering work of Caffarreli \cite{caffarelli1992regularity}, and the more recent results by De Philippis and Figalli \cite{de2013w} were established through the Monge-Amp\'{e}re equation.

\subsection{\bf Other approaches}

It is also worth pointing out the approach to optimal transport with quadratic cost using the \emph{semi-discrete} approximation. Namely several works \cite{AHA1998, merigot2011multiscale, kitagawa_merigot_thibert2016, levy2015} have considered the situation in which one of the measures considered has  density $d\mu = I_0 dx$, while the other is a sum of delta masses, $\mu = \sum q_i \delta_{y_i}$. 
It turns out that there exists weights $w_i$ such that the optimal transport map $x \mapsto y$ can be described via a power diagram. More precisely the set of $x$ mapping to $y_i$ is the following cell of the power diagram:
\[ PD_{w}(y_i) = \{x\::\: \textrm{ for all } j \quad  |x-y_i|^2 - w_i \leq |x-y_j|^2 - w_j\}. \]
The main observation (originating from \cite{AHA1998} and stated in \cite{merigot2011multiscale})
is that the weights $w_i$ are minimizers of the following unconstrained convex  functional
\[ F(w) = \sum_i \left(p_i w_i - \int_{PD_w(y_i)} \| x - y_i\|^2 - w_i d \mu(x)  \right). \]
 Works of Kitagawa, M\'erigot,  and Thibert \cite{ kitagawa_merigot_thibert2016}, Merigot \cite{merigot2011multiscale}, and Levy \cite{levy2015} use Newton based schemes and multiscale approaches to minimize the functional. The need to integrate over the power diagram makes the implementation somewhat geometrically delicate.  Nevertheless recent implementation by L\'evy
\cite{levy2015} gives impressive results in terms of speed. We also note that this approach provides the transportation mapping (not just the approximation of a plan). 
\medskip

Finally, we mention a recent work of Aude, Cuturi, Peyr\'e and Bach who investigated the possibility of using stochastic gradient descent on several formulations of OT problem with quadratic cost on large data sets. 

Other notable numerical techniques  that were  not covered here but could be valuable in a number of instances include \cite{mikami2008optimal,haber2010efficient,froese2013convergent, kitagawa2014, kuang2016, lindsey2016}.

\section{Applications}
\label{sec:applications}
In this section we review some recent applications of the optimal transport problem in signal and image processing, computer vision, and machine learning. 
\subsection{\bf Image retrieval}
\label{sec:image_retrieval}

 One of the earliest applications of the optimal transport problem was in image retrieval. Rubner, Tomasi, and Guibas \cite{rubner2000earth} employed the discrete Wasserstein metric, which they denoted as the Earth Mover's Distance (EMD), to measure the dissimilarity between image signatures.  In image retrieval applications, it is common practice to first extract features (i.e. color features, texture feature, shape features, etc.) and then generate high dimensional histograms or signatures (histograms with dynamic/adaptive binning), to represent images. The retrieval task then simplifies to finding images with similar representations (i.e. small distance between their histograms/signatures). 
 The Wasserstein metric is specifically suitable for such applications as it can compare histograms/signatures of different sizes (histograms with different binning). This unique capability turns the Wasserstein metric into an attractive candidate in image retrieval applications \cite{rubner2000earth,pele2009fast,li2013novel}. In \cite{rubner2000earth}, the Wasserstein metric was compared with common metrics such as the Jeffrey's divergence, the $\chi^2$ statistics, the $L_1$ distance, and the $L_2$ distance in an image retrieval task; and it was shown that the Wasserstein metric achieves the highest precision/recall performance amongst all. 
 
Speed of computation is an important practical consideration in image retrieval applications. For almost a decade, the high computational cost of the optimal transport problem overshadowed its practicality in large scale image retrieval applications. Recent advancements in numerical methods including the work of Pele and Werman \cite{pele2009fast}, Merigot \cite{merigot2011multiscale}, and Cuturi \cite{cuturi2013sinkhorn}, among many others, have reinvigorated optimal transport-based distances  as a feasible and appealing candidate for large scale image retrieval problems.

\subsection{\bf Registration and Morphing}
\label{sec:reg_morph}

Image registration deals with finding a common geometric reference frame between two or more images and it plays an important role in analyzing images obtained at different times or using different imaging modalities.  Image registration and more specifically biomedical image registration is an active research area.  Registration methods find a transformation $f$ that maximizes the similarity between two or more image representations (e.g. image intensities, image features, etc.). Comprehensive recent surveys on image registration can be found in 
\cite{sotiras2013deformable, crum2014non, oliveira2014medical}. Among the plethora of registration methods, nonrigid registration methods are especially important given their numerous applications in biomedical problems. They can be used to quantify the morphology of different organs, correct for physiological motion, and allow for comparison of image intensities in a fixed coordinate space (atlas). Generally speaking, nonrigid registration is a non-convex and non-symmetric problem, with no guarantee on existence of a globally optimal transformation.  

\begin{figure}[t!]
\includegraphics[width=\columnwidth]{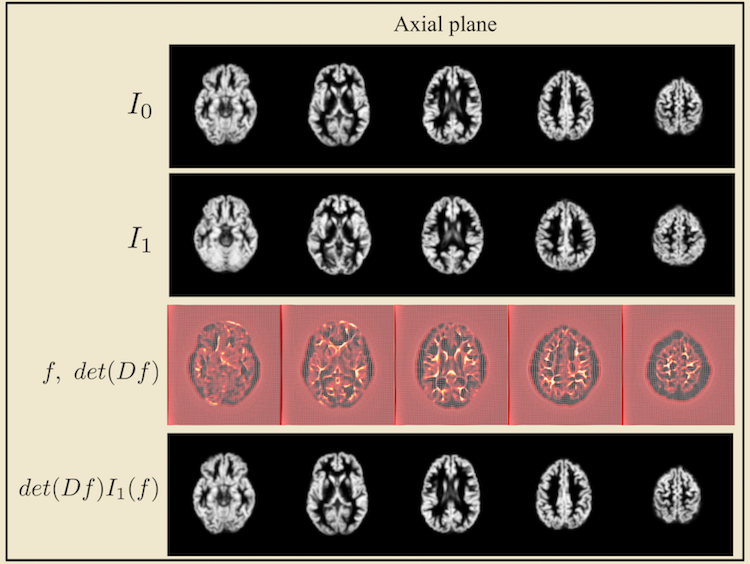}
\caption{An example of morphing a three dimensional image into another. The axial slices of images, and determinant of the Jacobian of the optimal transport map and the deformation  field is shown. The last row shows the slices of the morphed image.}
\label{fig:3DOT}
\end{figure}

 Haker et al. \cite{haker2001mass} proposed to use the Monge problem for image warping and elastic registration. Utilizing the Monge problem in an image warping/registration setting has a number of advantages. First, the existence and uniqueness of the global transformation (the optimal transport map) is known. Second, the problem is symmetric, meaning that the optimal transport map for warping $I_0$ to $I_1$ is the inverse of the optimal transport map for warping $I_1$ to $I_0$. Lastly, it provides a landmark-free and parameter-free registration scheme with a built-in mass preservation constraint. These advantages motivated several follow-up work to investigate the application of the Monge problem in image registration and warping \cite{haker2003monge,haker2004optimal,zhu2007image,ur20093d,museyko2009application,haber2010efficient}. Figure \ref{fig:3DOT} shows an example of morphing a  Magnetic Resonance Imaging (MRI) image of a brain into another. The axial slices of three dimensional images are shown, as well as slices of the determinant of the Jacobian of the optimal transport map and the deformation field.

 In addition to images, the optimal mass transport problem has also been used in point cloud and mesh registration \cite{makihara2010earth,flamary2014optimal,lai2014multi,su2015optimal}, which have various applications in shape analysis and graphics. In these applications, shape images (2D or 3D binary images) are first represented whether with sets of weighted points (i.e. point clouds), using clustering techniques such as K-means or Fuzzy C-means, or with meshes. Then, a regularized variation of the optimal transport problem is solved to match such representations. The regularization on the transportation problem is often imposed to enforce the neighboring points (or vertices) to remain near to each other after the transformation \cite{flamary2014optimal}. As another example, Su et al. \cite{su2015optimal} proposed a composition of conformal maps and optimal transport maps and introduced the {\it Conformal Wasserstein Shape Space}, which enables efficient shape matching and comparison.

\subsection{\bf Color transfer and texture synthesis}
\label{sec:color_texture}
Texture mixing and color transfer are among appealing applications of the optimal transport framework in image analysis, graphics, and computer vision. Here we briefly discuss these applications.

\subsubsection{\bf Color transfer}
\begin{figure}[t!]
\includegraphics[width=\columnwidth]{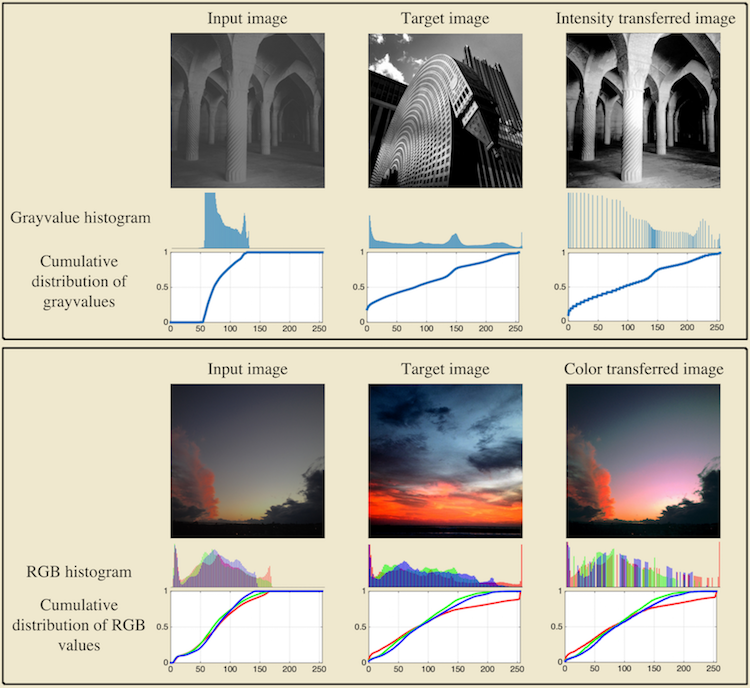}
\caption{Grayvalue and color transfer via optimal transportation.}
\label{fig:colortransfer}
\end{figure}

The purpose of color transfer \cite{reinhard2001color} is to change the color palette of an image to impose the feel and look of another image. The color transfer is generally performed through finding a map, which morphs the color distribution of the first image into the second one. For grayscale images, the color transfer problem simplifies to a histogram matching problem, which is solved through the one-dimensional optimal transport formulation \cite{delon2004midway}. In fact, the classic problem of histogram equalization is indeed a one-dimensional transport problem \cite{delon2004midway}. The color transfer problem, on the other hand, is concerned with  pushing the three-dimensional color distribution of the first image into the second one. This problem can also be formulated as an optimal transport problem as demonstrated in \cite{rabin2011wasserstein,rabin2014adaptive,ferradans2014regularized}. 

A complication that occurs in the color transfer on real images, however, is that a perfect match between color distributions of the images is often not satisfying. This is due to the fact that a color transfer map may not transfer the colors of neighboring pixels in a coherent manner, and may introduce artifacts in the color transferred image.  Therefore, the color transfer map is often regularized to make the transfer map spatially coherent \cite{rabin2014adaptive}. Figure \ref{fig:colortransfer} shows a simple example of grayvalue and color transfer via optimal transport framework. It can be seen that the cumulative distribution of the grayvalue and color transferred images are similar to that of the input image.

\subsubsection{\bf Texture synthesis and mixing}
Texture synthesis is the problem of synthesizing a texture image that is visually similar to an exemplar input texture image, and has various applications in computer graphics and image processing \cite{efros2001image,galerne2011random}. Many methods have been proposed for texture synthesis, among which are {\it synthesis by recopy} and {\it synthesis by statistical modeling}. Texture mixing, on the other hand, considers the problem of synthesizing a texture image from a collection of input texture images in a way that the synthesized texture provides a meaningful integration of the  colors and textures of the input texture images. Metamorphosis is one of the successful approaches in texture mixing, which performs the mixing via identifying correspondences between elementary features (i.e. textons) among input textures and progressively morphing between the shapes of elements. In other approaches, texture images are first parametrized through a tight frame (often steerable wavelets) and statistical modeling is performed on the parameters.   Rabin et al. \cite{rabin2012wasserstein}, for instance, used the optimal transport framework, specifically the sliced Wasserstein barycenters, on the coefficients of a steerable wavelet tight frame for texture mixing. 

\begin{figure}[t!]
\includegraphics[width=\columnwidth]{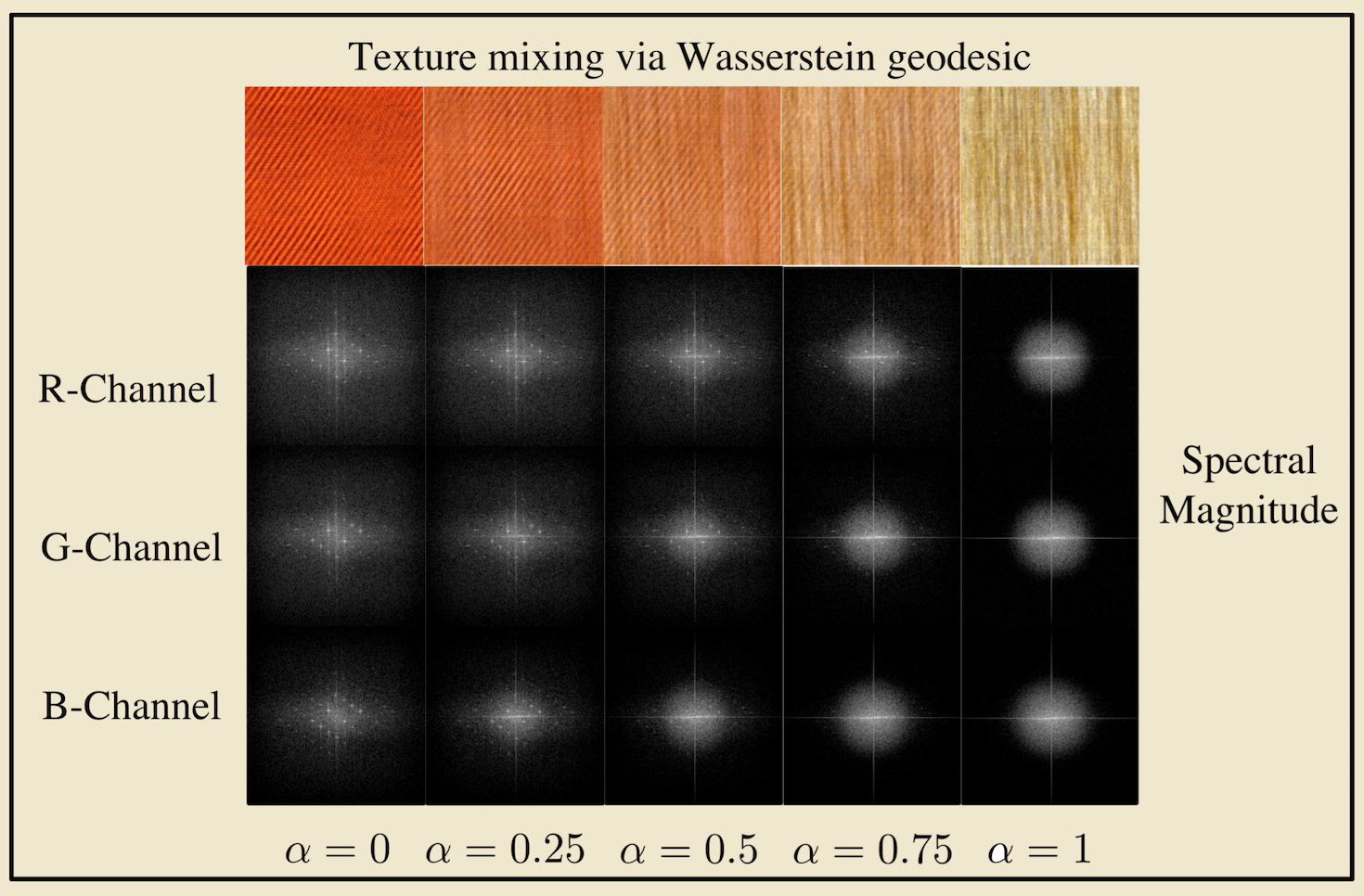}
\caption{An example of texture mixing via optimal transport using the method presented in Ferradans et al. \cite{ferradans2013static}}
\label{fig:texturemixing}
\end{figure}

 Other successful approaches include the random phase and spot noise texture modeling \cite{ferradans2013static}, which model textures as stationary Gaussian random fields \cite{galerne2011random}. Briefly, these methods are based on the assumption that the visual texture perception is based on the spectral magnitude of the texture image. Therefore, utilizing the spectral magnitude of an input image and randomizing its phase will lead to a new synthetic texture image which is visually similar to the input image. Ferradans et al. \cite{ferradans2013static} utilized this assumption together with the  Wasserstein geodesics to interpolate between spectral magnitude of texture images, and provide synthetic mixed texture images. Figure \ref{fig:texturemixing} shows an example of texture missing via the Wasserstein geodesic between the spectral magnitudes of the input texture images. The in-between images are synthetically generated using the random phase technique. 

\subsection{\bf Image denoising and restoration}
\label{sec:restoration}

The optimal transport problem has also been used in several image denoising and restoration problems \cite{lellmann2014imaging,rabin2011wasserstein, swoboda2013convex}. The goal in these applications is to restore or reconstruct an image from noisy or incomplete observation. Rabin and Peyr\'{e} \cite{rabin2011wasserstein}, for instance,  introduced an optimal transport-based framework to enforce statistical priors for constrained functional minimization. More precisely, they utilized the Sliced Wasserstein metric to measure the adequacy of the statistics of the reconstructed signal/image with the given constraints. In a similar approach, Swoboda and Sch\"{o}rr proposed a variational optimization problem with the Total Variation (TV) as the regularizer and the Wasserstein distance (dual formulation) to enforce prior statistics on reconstructed signal/image. They  relaxed the optimization into a convex/concave saddle point problem and solved it via a proximal gradient descent.

In another approach, Lellmann et al. \cite{lellmann2014imaging} utilized the Kantorovich-Rubinsten discrepancy term together with a Total Variation term in the context of image denoising. They called their method Kantorovich-Rubinstein-TV (KR-TV) denoising. It should be noted that, the Kantorovich-Rubinstein metric is closely related to the 1-Wasserstein metric (for one dimensional signals they are equivalent). The KR term in their proposed functional provides a fidelity term for denoising while the TV term enforces a piecewise constant reconstruction.

\subsection{\bf Transport based morphometry}
\label{sec:tbm}

Given their suitability for comparing mass distributions, transport-based approaches for performing pattern recognition of morphometry encoded in image intensity values have also recently emerged. Recently described approaches for transport-based morphometry (TBM) \cite{wang2013linear, basu2014detecting,kolouri2016continuous} work by computing transport maps or plans between a set of images and a reference or template image. The transport plans/maps are then utilized as an invertible feature/transform onto which pattern recognition algorithms such as principal component analysis (PCA) or linear discriminant analysis (LDA) can be applied. In effect, it utilizes the LOT framework described earlier in Section \ref{sec:LOT}. These techniques has been recently employed to decode differences in cell and nuclear morphology for drug screening \cite{basu2014detecting}, and cancer detection histopathology \cite{wang2011optimal,ozolek2014accurate} and cytology \cite{tosun2015detection}, amongst other applications including the analysis of galaxy morphologies \cite{kolouri2016continuous}, for example.

\begin{figure*}[t!]
\centering
\includegraphics[width=\linewidth]{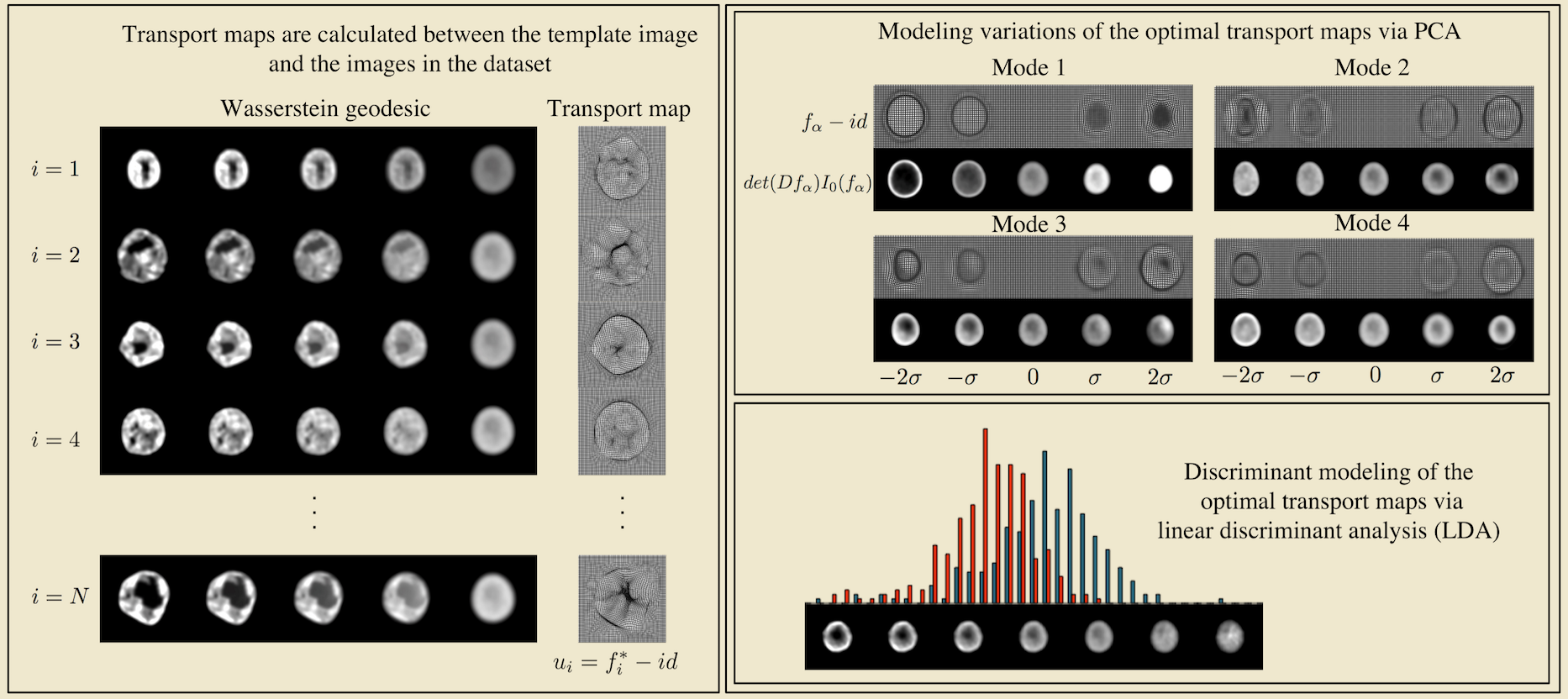}
\caption{The schematic of the TBM framework. The optimal transport maps between input images $I_1,...,I_N$ and a template image $I_0$ is calculated. Next, linear statistical modeling such as principal component analysis (PCA), linear discriminant analysis (LDA), and canonical correlation analysis (CCA) is performed on the optimal transport maps. The resulting transport maps obtained from the statistical modeling step are then applied to the template image to visualize the results of the analysis in the image space.}
\label{fig:TBM}
\end{figure*}

We note the strong similarity between deformation-based methods which have long been used to analyzed radiological images \cite{grenander1998computational,joshi2000landmark}, for example. The difference being that TBM allows for numerically exact, uniquely defined solutions for the transport plans or maps used. That is, images can be matched with little perceptible error. The same is not true in methods that rely on registration via the computation of deformations, given the significant topology differences commonly found in medical images. Moreover, TBM allows for comparison of the entire intensity information present in the images (shapes and textures), while deformation-based methods are usually employed to deal with shape differences. Figure \ref{fig:TBM} shows a schematic of the TBM steps applied to a cell nuclei dataset. It can be seen that the TBM is capable of modeling the variation in the dataset. In addition, it enables one to visualize the classifier, which discriminates between image classes (in this case malignant versus benign).

\subsection{\bf Super-Resolution}
\label{sec:super_resolution}

Super-resolution is the process of reconstructing a high-resolution image from one or several corresponding low-resolution images. Super-resolution algorithms can be broadly  categorized  into two major classes namely ``multi-frame'' super resolution and ``single-frame" super resolution,  based on the number of low-resolution images they require to reconstruct the corresponding high-resolution image. The transport-based morphometry approach was used for single frame super resolution in  \cite{kolouri2015transport} to reconstruct high-resolution faces from very low resolution input face images. The authors utilized the transport-based morphometry in combination with subspace learning techniques to learn a nonlinear model for the high-resolution face images in the training set.

\begin{figure*}[t!]
\centering
\includegraphics[width=\linewidth]{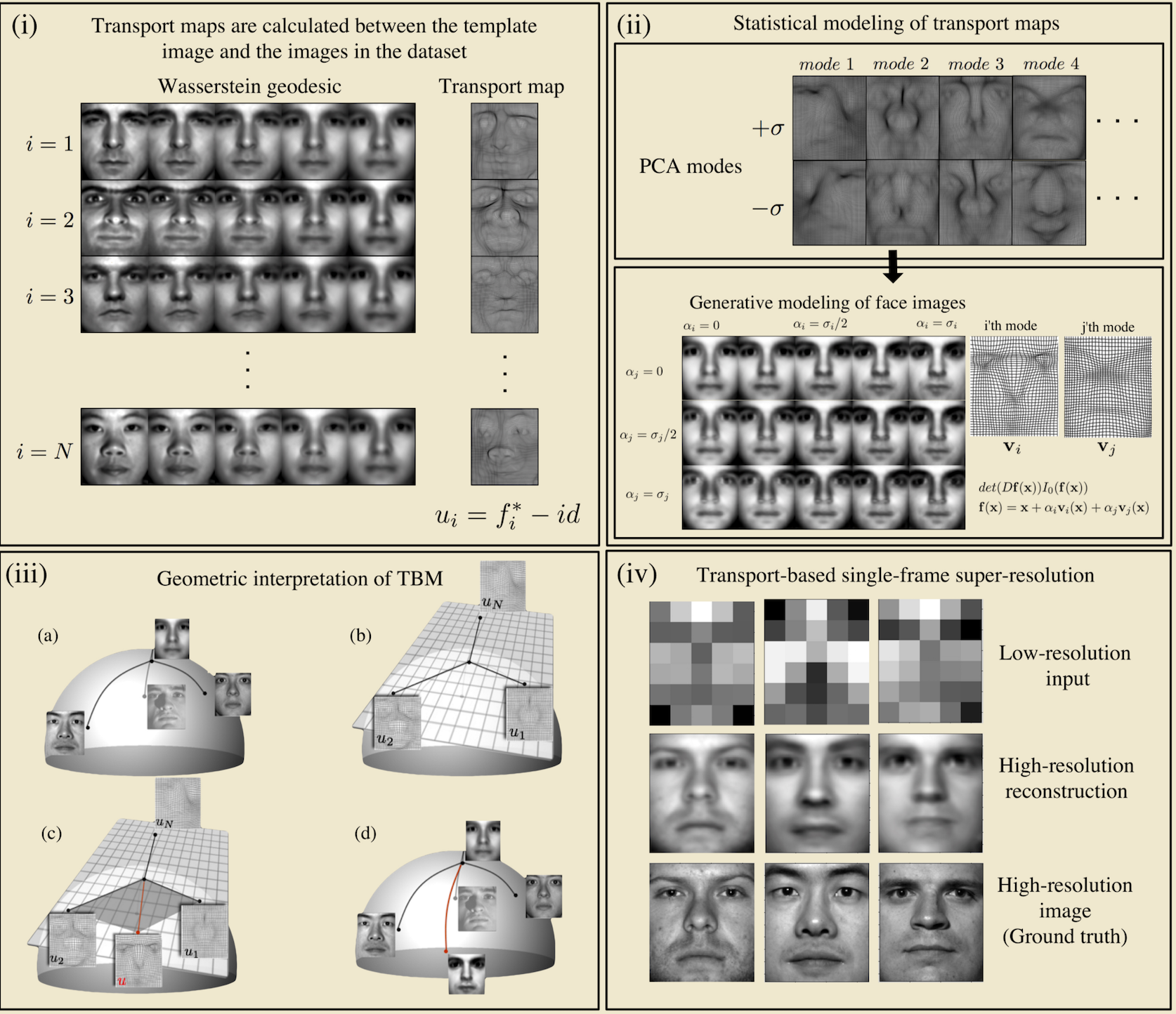}
\caption{In the training phase, optimal transport maps that morph the template image to high-resolution training face images are calculated (i). Principal component analysis (PCA) is used to learn a linear subspace for transport maps for which a linear combination of obtained eigenmaps can be applied to the template image to obtain synthetic face images  (ii). A geometric interpretation of the problem is depicted in panel (iii), and reconstruction results are shown in panel (iv).}
\label{fig:TBSFSR}
\end{figure*}

In short, the method consists of a training and a testing phase. In the training phase, it uses high resolution face images and morphs them to a template high-resolution face through optimal transport maps. Next, it learns a subspace for the calculated optimal transport maps. A transport map in this subspace can then be applied to the template image to synthesize a high-resolution face image. In the testing phase, the goal is to reconstruct a high-resolution image from the low-resolution input image. The method searches for a synthetic high-resolution face image (generated from the transport subspace) that provides a corresponding low-resolution image which is similar to the input low-resolution image. Figure \ref{fig:TBSFSR} shows the steps used in this method and demonstrates reconstruction results. 

\subsection{\bf Machine-Learning and Statistics}
\label{sec:machine_learning}

The optimal transport framework has recently attracted ample attention from the machine learning and statistics communities \cite{solomon2014wasserstein,courty2014domain,frogner2015learning,kolouri2016radon,kolouri2016sliced,montavon2015wasserstein}. Some applications of the optimal transport in these arenas include various transport-based learning methods \cite{montavon2015wasserstein,frogner2015learning,kolouri2016sliced,solomon2014wasserstein}, domain adaptation, Bayesian inference \cite{flamary2014optimal,courty2014domain,courty2015optimal}, and hypothesis testing \cite{del1999tests,ramdas2015wasserstein} among others. Here we provide a brief overview of the recent developments of transport-based methods in machine learning and statistics.

\subsubsection{\bf Learning} Transport-based distances have been recently used in several works as a loss function for regression, classification, etc. Montavon, M\"{u}ller, and Cuturi \cite{montavon2015wasserstein}  for instance utilized the dual formulation of the entropy regularized Wasserstein distance to train restricted Boltzmann machines (RBMs). Boltzmann machines are probabilistic graphical models (Markov random fields) that can be categorized as stochastic neural networks and are capable of extracting hierarchical features at multiple scales. RBMs are bipartite graphs which are special cases of Boltzmann machines, which define parameterized probability distributions over a set of $d$-binary input variables (observations) whose states are represented by $h$ binary output variables (hidden variables).  RBMs' parameters are often learned through information theoretic divergences such as KL-Divergence. Montavon et al. \cite{montavon2015wasserstein} proposed an alternative approach through a scalable entropy regularized Wasserstein distance estimator for RBMs, and showed the practical advantages of this distance over the commonly used information divergence-based loss functions. 

In another approach, Frogner et al. \cite{frogner2015learning} used the entropy regularized Wasserstein loss for multi-label classification. They proposed a relaxation of the transport problem to deal with unnormalized measures by replacing the equality constraints in Equation \eqref{eq:lp} with soft penalties with respect to KL-divergence. In addition, Frogner et al. \cite{frogner2015learning} provided statistical bounds on the expected semantic distance between the prediction and the groundtruth. In yet another approach, Kolouri et al. \cite{kolouri2016sliced} utilized the sliced Wasserstein metric and provided a family of positive definite kernels, denoted as Sliced-Wasserstein Kernels, and showed the advantages of learning with such kernels. The Sliced-Wasserstein Kernels were shown to be effective in various machine learning tasks including classification, clustering, and regression.  

Solomon et al. \cite{solomon2014wasserstein} considered the problem of graph-based semi-supervised learning, in which graph nodes are partially labeled and the task is to propagate the labels throughout the nodes. Specifically, they considered a problem in which the labels are histograms. This problem arises for example in traffic density prediction, in which the traffic density is observed for few stop lights over 24 hours in a city and the city is interested in predicting the traffic density in the un-observed stop lights. They pose the problem as an optimization of a Dirichlet energy for distribution-valued maps based on the 2-Wasserstein distance, and present a Wasserstein propagation scheme for semi-supervised distribution propagation along graphs.

\subsubsection{\bf Domain adaptation } Domain adaptation is one of the fundamental problems in machine learning which has gained proper attention from the machine learning research community in the past decade \cite{patel2015visual}. Domain adaptation is the task of transferring knowledge from classifiers trained on available labeled data to unlabeled test domains with data distributions that differ from that of the training data. 
The optimal transport framework is recently presented as a potential major player in domain adaptation problems \cite{flamary2014optimal,courty2014domain,courty2015optimal}. 
Courty, Flamary, and Davis \cite{courty2014domain}, for instance, assumed that there exists a non-rigid transformation between the source and target distributions and find this transformation using an entropy regularized optimal transport problem.  They also proposed a label-aware version of the problem in which the transport plan is regularized so a given target point (testing exemplar) is only associated with source points (training exemplars) belonging to the same class. Courty et al. \cite{courty2014domain} showed that domain adaptation via regularized optimal transport outperform the state-of-the-art results in several challenging domain adaptation problems. 

\subsubsection{\bf Bayesian inference} 
\begin{figure}
\centering
\includegraphics[width=\columnwidth]{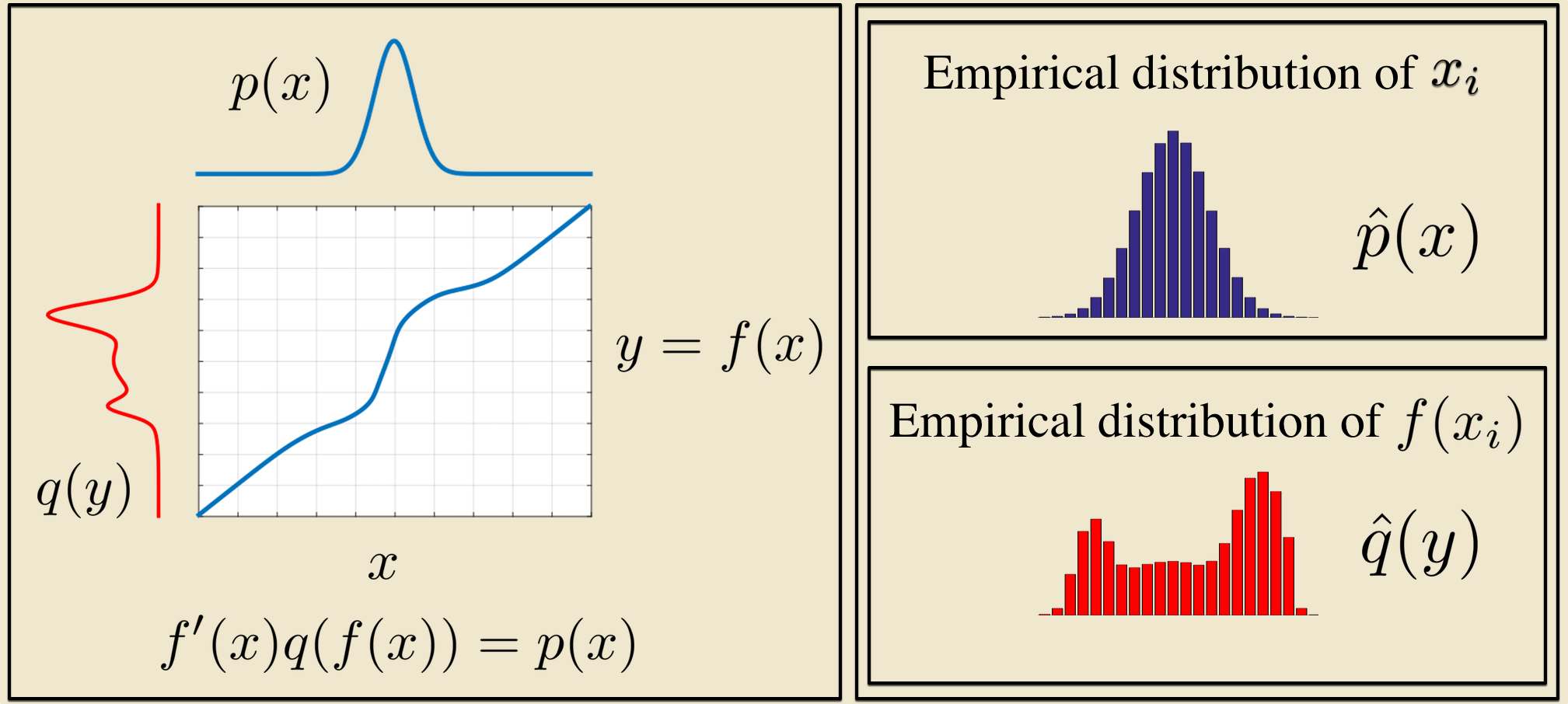}
\caption{ Left panel shows the prior distribution $p$ and the posterior distribution $q$ and the corresponding transport map $f$ that pushes $p$ into $q$. One million samples, $x_i$, were generated from distribution $p$ and the top-right panel shows the empirical distribution of these samples denoted as $\hat{p}$. The bottom-right panel shows the empirical distribution of transformed samples, $y_i=f(x_i)$, denoted as $\hat{q}$. }
\label{fig:sampling}
\end{figure}

Another interesting and emerging application of the optimal transport problem is in Bayesian inference \cite{el2012bayesian,kim2013efficient,reich2013nonparametric,oliver2014minimization}. In Bayesian inference, one critical step is the evaluation of expectations with respect to a posterior probability function, which leads to complex multidimensional integrals. These integrals are commonly solved through the Monte Carlo numerical integration, which requires independent sampling from the posterior distribution. In practice, sampling from a general posterior distribution might be difficult, therefore the sampling is performed via a Markov Chain which converges to the posterior probability after certain number of steps. This leads to the celebrated Markov Chain Monte Carlo (MCMC) method. The downside of MCMC is that the samples are not independent and hence the convergence of the empirical expectation is slow.  El Moselhy and Marzouk \cite{el2012bayesian} proposed a transport-based method that evades the need for Markov chain simulation by allowing direct sampling from the posterior distribution. The core idea in their work is to find a transport map (via a regularized Monge formulation), which pushes forward the prior measure to the posterior measure. Then, sampling the prior distribution and applying the transport map to the samples, will lead to a sampling scheme from the posterior distribution. Similar ideas were used in \cite{kim2013efficient,oliver2014minimization}. Figure \ref{fig:sampling} shows the basic idea behind these methods. 

\subsubsection{\bf Hypothesis testing}  Wasserstein distance is used for goodness of fit testing in \cite{del1999tests} and for two sample testing in \cite{ramdas2015wasserstein}. Ramdas et al. \cite{ramdas2015wasserstein} presented connections between the entropy regularized Wasserstein distance, multivariate Energy distance, and the kernel maximum mean discrepancy (MMD) \cite{gretton2012kernel}, and provided a ``distribution free'' univariate Wasserstein test statistic. These and other applications of transport-related concepts show the promise of the mathematical modeling technique in the design of statistical data analysis methods to tackle modern learning problems.

Finally, we note that in the interest of brevity, a number of other important applications of transport-related techniques were not discussed above, but are certainly interesting on their own right. These include astronomical sciences \cite{frisch2002reconstruction,brenier2003reconstruction, frisch2004application}, meteorology sciences \cite{budd2013monge}, optics \cite{graf2012optimal,bauerle2012algorithm}, particle image velocimetry \cite{saumier2015optimal}, information theory \cite{tannenbaumCDC},  optical flow estimation \cite{kolesov2010fire,kato2015optical}, and geophysics \cite{metivier2016measuring,engquist2016optimal} among others.

\section{Summary and Conclusions}

Transport-related methods and applications have come a long way. While earlier applications focused primarily in civil engineering and economics problems, they have recently begun to be employed in a wide variety of problems related to signal and image analysis, and pattern recognition. In this tutorial, seven main areas of application were reviewed: image retrieval \ref{sec:image_retrieval}, registration and morphing \ref{sec:reg_morph}, color transfer and texture analysis \ref{sec:color_texture}, image restoration \ref{sec:restoration}, transport-based morphometry \ref{sec:tbm}, image super-resolution \ref{sec:super_resolution}, and machine learning and statistics \ref{sec:machine_learning}.  As evidenced by the number of papers published (see Fig \ref{fig:OTRef}) per year, transport and related techniques have received increased attention in recent years. Overall, researchers have found that the application of transport-related concepts can be helpful to solve problems in diverse applications. Given recent trends, it seems safe to expect that the the number of different application areas will continue to grow.

In its most general form, the transport-related techniques reviewed in this tutorial can be thought as mathematical models for signals, images, and in general data distributions. Transport-related metrics involve calculating differences not only of pixel or distribution intensities, but also "where" they are located in the corresponding coordinate space (a pixel coordinate in an image, or a particular axis in some arbitrary feature space). As such, the geometry (e.g. geodesics) induced by such metrics can give rise to dramatically different algorithms and data interpretation results. The interesting performance improvements recently obtained could motivate the search for a more rigorous mathematical understanding of transport-related metrics and applications.

We note that the emergence of numerically precise and efficient ways of computing transport-related metrics and geodesics, presented in section \ref{sec:numerics} also serves as an enabling mechanism. Coupled with the fact that several mathematical properties of transport-based metrics have been extensively studied, we believe that the ground is set of their increased use as foundational tools or building blocks based on which complex computational systems can be built. The confluence of these emerging ideas may spur a significant amount of innovation in a world where sensor and other data is becoming abundant, and computational intelligence to analyze these is in high demand. We believe transport-based models while become an important component of the ever expanding tool set available to modern signal processing and data science experts.

\section{Acknowledgements}

Authors gratefully acknowledge funding from the NSF (CCF 1421502) and the NIH (GM090033, CA188938) in contributing to a portion of this work. DS also acknowledges funding by NSF (DMS DMS-1516677)

\bibliographystyle{IEEE}
\bibliography{OTReview_Arxiv}

\begin{IEEEbiography}[{\includegraphics[width=1in,height=1.25in,clip,keepaspectratio]{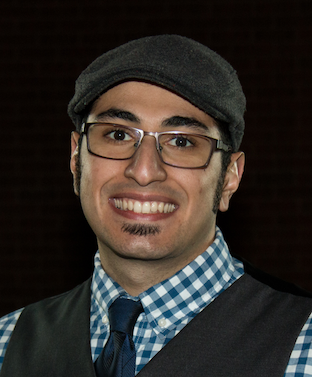}}]%
{Soheil Kolouri}
received his B.S. degree in electrical engineering from Sharif University of Technology, Tehran, Iran, in 2010, and his M.S. degree in electrical engineering in 2012 from Colorado State University, Fort Collins, Colorado. He earned his doctorate degree in biomedical engineering from Carnegie Mellon University in 2015. His thesis, titled, ``Transport-based pattern recognition and image modeling'', won the best thesis award. He is currently at HRL Laboratories, Malibu, California, United States. 
\end{IEEEbiography}

\begin{IEEEbiography}[{\includegraphics[width=1in,height=1.25in,clip,keepaspectratio]{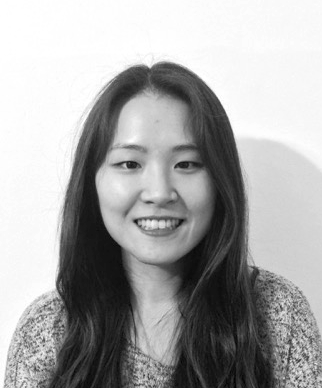}}]%
{Serim Park} received her B.S. degree in Electrical and Electronic Engineering from Yonsei University, Seoul, Korea in 2011 and is currently a doctoral candidate in the Electrical and Computer Engineering Department at Carnegie Mellon University, Pittsburgh, United States. She is mainly interested in signal processing and machine learning, especially designing new  signal and image transforms and developing novel systems for pattern recognition.
\end{IEEEbiography}

\begin{IEEEbiography}[{\includegraphics[width=1in,height=1.25in,clip,keepaspectratio]{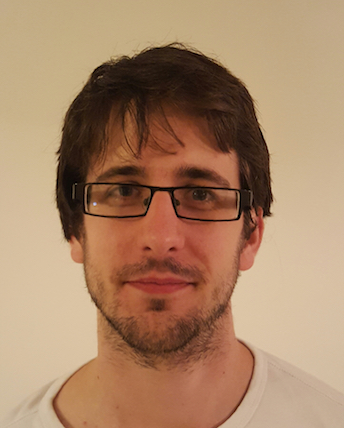}}]%
{Matthew Thorpe} received his BSc, MSc and PhD in Mathematics from the University of  Warwick, UK in 2009, 2012 and 2015 respectively and his MScTech in Mathematics from the University of New South Wales, Australia, in 2010. He is currently a postdoctoral associate within the mathematics department at Carnegie Mellon University.
\end{IEEEbiography}

\begin{IEEEbiography}[{\includegraphics[width=1in,height=1.25in,clip,keepaspectratio]{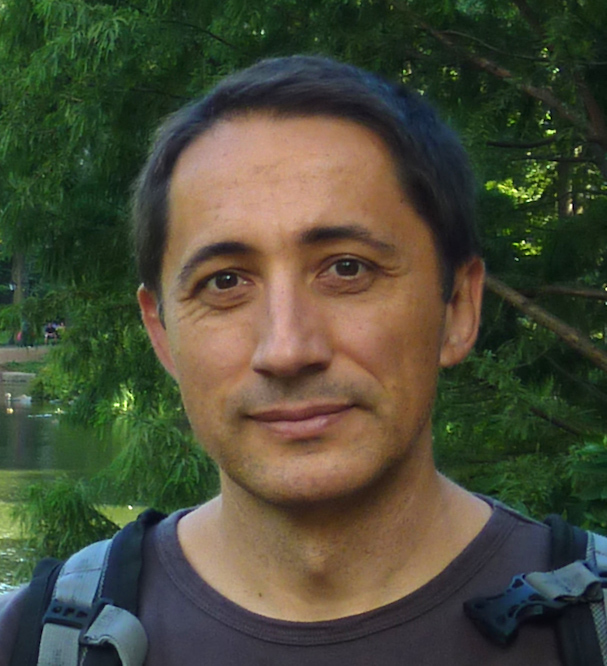}}]%
{Dejan Slep\v{c}ev}
earned B.S degree in mathematics from the University of Novi Sad in 1995, M.A. degree in mathematics from University of Wisconsin Madison in 2000 and Ph.D. in mathematics from University of Texas at Austin in 2002. He is currently associate professor at the Department of Mathematical Sciences at Carnegie Mellon University.
\end{IEEEbiography}

\begin{IEEEbiography}[{\includegraphics[width=1in,height=1.25in,clip,keepaspectratio]{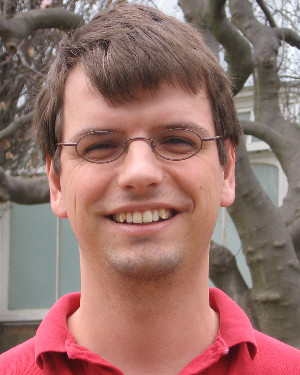}}]%
{Gustavo K. Rohde}
earned B.S. degrees in physics and mathematics in 1999, and the M.S. degree in electrical engineering in 2001 from Vanderbilt University. He received a doctorate in applied mathematics and scientific computation in 2005 from the University of Maryland. He is currently an associate professor of Biomedical Engineering, and Electrical and Computer Engineering at the University of Virginia. Contact: gustavo@virginia.edu.
\end{IEEEbiography}

\end{document}